\documentclass[preprint,12pt]{elsarticle}
\usepackage{booktabs} 
\usepackage{adjustbox}

\usepackage{amssymb}
\usepackage{listings}
\usepackage{hyperref}
\usepackage{xcolor}

\usepackage{siunitx}
\usepackage{subcaption}
\usepackage{caption}
\usepackage{graphicx}

\usepackage{multirow}

\usepackage{soul}

\newtoggle{finalPaper}
\toggletrue{finalPaper} 

\iftoggle{finalPaper} {
	\newcommand{\addtxt}[1]{#1}
	
	\newcommand{\rmvtxt}[1]{}}
{
	\newcommand{\addtxt}[1]{\textcolor{blue}{#1}}
	
	\newcommand{\rmvtxt}[1]{\st{#1}}}

\begin{document}
\begin{frontmatter}

\author[inst1]{David de-Fitero-Dominguez}
\author[inst2]{\corref{cor1}Mariano Albaladejo-González}
\ead{mariano.albaladejog@um.es}

\author[inst1]{Antonio Garcia-Cabot}

\author[inst1]{Eva Garcia-Lopez}

\author[inst1]{Antonio Moreno-Cediel}

\author[inst3]{Erin Barno}

\author[inst3]{Justin Reich}

\affiliation[inst1]{
            organization={\addtxt{Departamento de Ciencias de la Computación,} Universidad de Alcalá},
            addressline={Campus Universitario. Ctra. Madrid-Barcelona, Km. 33,600}, 
            city={Alcalá de Henares},
            postcode={28805}, 
            state={Madrid},
            country={Spain}}

\affiliation[inst2]{
            organization=\addtxt{{Departamento de Ingeniería de la Información y de las Comunicaciones,} Universidad de Murcia},
            addressline={Calle Campus Universitario}, 
            city={Murcia},
            postcode={30100}, 
            state={Murcia},
            country={Spain}}

\affiliation[inst3]{
            organization= \addtxt{{Comparative Media Studies/Writing,} Massachusetts Institute of Technology},
            addressline={Cambridge}, 
            postcode={MA 02139}, 
            state={Massachusetts},
            country={USA}}
\cortext[cor1]{Corresponding author}
\title{Evaluating Large Language Models for automatic analysis of teacher simulations}

\begin{abstract}
Digital Simulations (DS) provide safe environments where users interact with an agent through conversational prompts, providing engaging learning experiences that can be used to train teacher candidates in realistic classroom scenarios. These simulations usually include open-ended questions, allowing teacher candidates to express their thoughts but complicating an automatic response analysis. To address this issue, we have evaluated Large Language Models (LLMs) to identify characteristics (user behaviors) in the responses of DS for teacher education. We evaluated the performance of DeBERTaV3 and Llama 3, combined with zero-shot, few-shot, and fine-tuning. Our experiments discovered a significant variation in the LLMs' performance depending on the characteristic to identify. Similar variations also occurred in Phi-4-mini and Qwen-3. Additionally, we noted that DeBERTaV3 significantly reduced its performance when it had to identify new characteristics. In contrast, Llama 3 performed better than DeBERTaV3 in detecting new characteristics and showing more stable performance. Therefore, in DS where teacher educators need to introduce new characteristics because they change depending on the simulation or the educational objectives, it is more recommended to use Llama 3. These results can guide other researchers in introducing LLMs to provide automatic evaluations in DS.
\end{abstract}


\end{frontmatter}


\section{Introduction}
\label{sec:introduction}


In recent years, we have witnessed significant technological growth that has impacted numerous sectors, including education \cite{timotheou2023impacts}. One noteworthy example is Digital Simulations (DS), which can be used to improve learning experiences \cite{Ledger2022}. The quality of education is intrinsically linked to the competence and preparation of those who deliver it, so it is essential to develop simulators to support teacher education \cite{BasilottaGmezPablos2022}. For this reason, Teacher Moments (TM), a web application to provide DS, has been developed for teaching practice. The DS are meticulously scripted simulations where participants interact with an agent through conversational prompts, creating a dynamic and interactive learning experience \cite{LittenbergTobias2022}. Through TM, teacher candidates can practice real classroom scenarios in a controlled and safe environment. TM allows teacher candidates to practice classroom scenarios without real students, which prevents their mistakes from negatively impacting students' education \cite{Ledger2022}. Since TM is accessible on any computer connected to the Internet, at any time, and from any location, it provides a flexible platform for continuous teaching development. Teacher candidates can repeatedly simulate teaching experiences, which is one of the most effective methods for honing practical teaching skills to manage diverse classroom situations, respond to unexpected student behaviors, and integrate effective teaching strategies \cite{Reich2022}.

It is important to complement educational simulations with an evaluation of the users' responses and provide them with personalized feedback \cite{MAIER2022100080}. Without this feedback, teacher candidates could continuously reproduce wrong behaviors, leading to the internalization of bad habits \cite{cavalvanti2021}. Therefore, in DS, it is essential to analyze the users' responses and provide them with appropriate feedback \cite{DEEVA2021104094}. However, manually analyzing the users' responses brings scalability issues and limits the flexibility of the simulators. An alternative is to automatically analyze the users' responses using Artificial Intelligence (AI) \cite{Kurni2023}.

In DS, it is common to provide open-ended questions, allowing teacher candidates to express their thoughts and beliefs \cite{LittenbergTobias2022}. However, this method complicates the automatic analysis of responses, since it demands an in-depth understanding to identify specific behaviors or characteristics. From this section onward, we define a characteristic as a behavior that an educator needs to identify in a user's response to provide appropriate feedback. For instance, recognizing unacceptable behaviors such as rudeness or unfairness towards a student. Identifying these educational characteristics is even more complex because the characteristics to be identified can also be open-ended and not predefined. Teacher educators can create different simulations based on the competencies they want to train or evaluate. Teacher candidates complete the simulations, and the feedback they should receive for their responses depends on the circumstances and the educational goals set by teacher educators in the simulation \cite{LittenbergTobias2022}. Consequently, we face a classification problem where the element to classify is variable. Due to the traditional supervised AI model's failure to generalize to new contexts and problems, they are not suitable for identifying variable characteristics in users' responses.

Large language models (LLMs) have shown significant performance improvements and capabilities across various domains in the last few years \cite{Chang2024}. Due to their pre-training, they outperform traditional supervised AI models in natural language processing tasks \cite{min2023recent}. As a result, LLMs are the most suitable for automatically analyzing responses in DS. However, it is necessary to evaluate any AI model before integrating it into real-world applications \cite{Naser2021}. Consequently, before integrating LLMs in DS and any AI model in education, it is crucial to evaluate their behavior to identify educational characteristics, especially how they perform with unseen characteristics that an educator could introduce. To the best of our knowledge, no authors have previously conducted this analysis that can guide researchers to integrate LLMs to provide personalized feedback in DS.

To address this gap, we have evaluated whether the performance of LLMs depends on the educational characteristics to identify. It is important to know whether the LLM's capability to identify specific behaviors in DS varies depending on the behavior to identify. We have also analyzed how zero-shot, few-shot, and fine-tuning impact the model's performance in previously seen characteristics and in new (unseen) characteristics. This second analysis reports the performance of LLMs for analyzing DS responses in the case where we have predefined behaviors and where a teacher educator needs to analyze a new behavior. To develop these analyses, we have employed a dataset with users' responses in DS for teacher education in TM. Expert educators have manually identified and labeled 14 characteristics, with a total of 4,822 response/characteristic pairs. These experiments aim to answer the following Research Questions (RQs) that should be considered when introducing LLMs to provide feedback in DS:

\begin{itemize}

\item \textbf{RQ1}. Does the effectiveness of LLMs in identifying educational characteristics within DS for teacher education change depending on the characteristic itself?

\item \textbf{RQ2}. What is the best LLM configuration to integrate into DS for teacher education when the characteristics to identify are predefined?

\item \textbf{RQ3}. What is the best LLM configuration to integrate into DS for teacher education when the characteristics to identify are open-ended?

\end{itemize}

The rest of the paper is organized as follows. Section \ref{sec:related_works} provides a background of LLMs in education. Section \ref{sec:methodology} includes the methodology followed to answer the aforementioned RQs. Section \ref{sec:results} shows the results obtained, and Section \ref{sec:discussion} contains the discussion about the outcomes. Finally, we present the research conclusions and future work in Section \ref{sec:conclusions}.

\section{Literature review}
\label{sec:related_works}
This section examines the literature on the use of LLMs in educational settings, including their application for automatic response analysis.

\subsection{LLMs in education}
Advancements in LLMs have driven the development of numerous educational technology innovations \cite{Yan2023}. These innovations aim to automate time-consuming and laborious tasks of generating and analyzing data, typically text \cite{tunstall2022natural}. LLMs are based on the transformer architecture, which in its original design consisted of an encoder and/or a decoder \cite{Patwardhan2023}. Bidirectional Encoder Representations from Transformers (BERT) \cite{Shreyashree2022} employs an encoder-only architecture. In contrast, other models use a decoder-only architecture, such as Llama 3  \citep{llama3modelcard}. Nevertheless, both types of LLMs are pre-trained on gigantic text datasets, reducing the data necessary to achieve the targeted task of the model \cite{Yenduri2024}. After this pre-training, there are three types of paradigms to use these models \cite{Patwardhan2023}:

\begin{itemize}

\item \textbf{Zero-shot}. The model does not receive any additional training specific to the task to be achieved. In this case, the model only has the knowledge acquired during the pre-training. 

\item \textbf{Few-shot}. The model receives a few examples of the task to achieve. These additional examples are usually provided to the model through the prompt inserted into the LLMs.

\item \textbf{Fine-tuning}. The model undergoes further training with a dataset related to the specific task to achieve. 
\end{itemize}

Due to their comprehensive architectures and large pre-training, several researchers have applied LLMs to educational settings. Personalized learning is one of the key applications, focused on adapting the learning experience to the student's particular needs \cite{Sharma2025}. \citet{Wang2025} proposed LearnMate, an LLM-based system that generates personalized learning plans and learning support. Similarly, \citet{Chen2024} developed GPTutor, a personalized tutor based on GPT-4 that adapts educational content and practice exercises to the student's interests and career goals. \citet{Hu2025} proposed an innovative approach to utilize GPT-4 to enhance the quality of teaching plans. The LLM simulates teacher-student interactions, generates teaching reflections, and uses them to refine the teaching plan. Another application of LLMs in education is sentiment analysis, especially in online learning environments, where educators lack direct visual contact with students. \citet{Hamdi2025} introduced a sentiment-based analysis of student engagement using LLMs for e-learning platforms. In addition, \citet{Shaikh2023} classified students' feedback into positive, negative, or neutral sentiments, achieving an F1 score of 0.88 using GPT-3.5.

\subsection{LLMs for automatic response analysis}
A valuable application of LLMs in education is the analysis of student responses. This is particularly beneficial in large-scale educational settings where manual grading would be impractical \cite{app11010095}. Additionally, LLMs can offer more consistent and unbiased evaluations of students. \citet{mizumoto2023exploring} employed GPT-3 text-davinci-003 for automatic essay assessment. Similarly, \cite{Mendona2025} evaluated Llama 3.2 and GPT-4o for automated scoring in formative assessments. Another example is \cite{10673924}, where the authors used GPT-3.5, GPT-4, and Gemini-Pro for an AI-based exam grader. \citet{Morris2024} applied DeBERTa to score math responses, a challenging task due to the logic and reasoning component of the responses. It is worth mentioning that \citet{Dai2023} observed an irregular performance of GPT-3.5 in zero-shot tasks for evaluating student essays. The precision and recall in identifying essays that did not follow the topic established by the educator were zero, which indicates the need to thoroughly evaluate these models before using them in real-world environments.

Our experiments also focus on the automatic analysis of responses, but based on the identification of specific characteristics in the participants' responses. The final output in our application is not an overall grade, as in the aforementioned works. Instead, we focus on detecting the presence or absence of specific characteristics in the participants' responses. Based on this, educators can establish personalized feedback. With the aim of identifying specific characteristics, \citet{gombert2023coding} analyzed free-text responses of German K–12 students to detect energy-related features during formative science assessments. The authors evaluated LLMs and traditional supervised AI models for this task. In our experiments, the characteristics are specifically related to teacher conduct, and the responses are collected through simulations rather than exams or assignments. Additionally, the LLMs utilized are BERT-based models, overlooking important decoder-based LLMs such as Llama 3. Prompt engineering techniques, such as zero-shot and few-shot evaluation, were also not explored, as the models used in that study did not support such approaches.

The proposed automatic evaluation could be complemented with expert knowledge to provide feedback to the students. Educators can determine what comments and recommendations to show based on the presence or absence of specific characteristics, similar to a rule-based system \cite{Ochoa2020}. This approach is particularly valuable as it enables teacher educators to provide feedback aligned with the educational objectives and scope of the developed simulation, effectively combining automated assessment with expert insights \cite{DEEVA2021104094}. For instance, a teacher educator could analyze whether teacher candidates are reprimanding a child in front of their classmates and, in such cases, offer recommendations on why this practice should be avoided and how to handle such situations properly.

LLMs have also been applied to improve professionals' training and education \cite{Sallam2023}. For instance, in medical education, they have been used for personalized learning, automatic assessments, writing assistance, and literature reviews \cite{Abdalrazaq2023}. In the context of teacher training, \citet{Bhowmik2024} utilized GPT-3.5 to create a virtual student conversation agent. In addition, \citet{Zheng2025} evaluated a fine-tuned Chinese version of Llama for simulating students in virtual trainings for teachers. The authors concluded that the LLM created realistic and engaging environments, significantly benefiting teaching skill development. Nevertheless, these works did not address the identification of characteristics in users' responses. To the best of our knowledge, only one previous study has focused on identifying educational characteristics in DS designed for the training of professionals \cite{LittenbergTobias2022}. This study used a smaller version of our dataset and tested supervised AI models and GPT-3 for characteristic identification. However, the study did not assess whether the models' performance varied across different characteristics. Furthermore, it did not evaluate the models' performance on new, unseen characteristics that an educator might introduce. This previous work also did not fine-tune the LLMs. Thus, there is a need to provide guidelines on these aspects for future researchers before they integrate LLMs into their DS.


\section{Material and methods}
\label{sec:methodology}
To address the research questions, a dataset including teacher candidates' responses collected from the TM platform was used. These responses were meticulously labeled by experts to identify specific educational characteristics. To evaluate the effectiveness of various LLMs in identifying these characteristics within DS, different configurations were implemented and tested on different subsets of the data. The goal was to assess how well the models could identify certain characteristics in teacher candidates' responses.

Specifically, the LLMs Llama 3 \citep{llama3modelcard} and DeBERTaV3 \citep{he2021debertav3} were employed. Llama 3 was used in zero-shot, few-shot, and fine-tuning configurations. Complementing this, DeBERTaV3, an encoder-only architecture known for its efficacy in classification tasks \citep{tunstall2022natural,he2021debertav3}, was also evaluated. The assessment focused on the models' ability to identify both characteristics present in the training data and new ones. This comprehensive approach served a dual purpose: to determine the most effective LLM configurations for predefined characteristic identification tasks, and to explore their potential in identifying open-ended characteristics within the teacher candidates' responses.

In the following sections, the methodology and experimental setup will be detailed. This includes the process of data collection and labeling, the specific configurations, and the strategic division of the dataset into subsets for training and evaluation purposes. The evaluation metrics used to assess the performance of the models will also be described. Additionally, the results of using Llama 3 in zero-shot, few-shot, and fine-tuning configurations, as well as DeBERTaV3's, will be presented and analyzed.

\subsection{Dataset compilation}
The dataset was collected from a simulation known as Jeremy's Journal. In this simulation, participants assume the role of a middle school English teacher who assists a student named Jeremy Green, who is facing personal challenges. This simulation was integrated into an online professional learning course for teacher candidates, conducted in early 2021. The course was attended by 5,458 participants, but the focus was placed on the subset of participants who completed the simulation and consented to participate in the research, totaling 494 participants. 

In this study, the responses to three specific questions about the case presented in the simulation were analyzed. To evaluate these responses, fourteen distinct characteristics were identified, each represented by a binary label (1 if the characteristic is present, 0 if it is not). A group of teacher education experts defined the labels by considering both their relevance for analyzing novice teachers’ classroom behavior and how clearly each could be recognized within the specific situations presented in Jeremy’s Journal simulation. Table \ref{tab:characteristics} describes the characteristics labeled. Not all characteristics were labeled in every vignette. In some vignettes, the situation depicted did not elicit user responses related to a given characteristic, making it difficult to identify the corresponding label. This process aimed to provide more accurate labeling of characteristics through responses. The labeling process was conducted by three expert evaluators. The binary labels they assigned constitute the ground truth required for fine-tuning and the performance of the language models.

\begin{table}[htbp]
    \centering
    \begin{tabular}{lp{9.5cm}}
    \toprule
    \textbf{Characteristic} & \textbf{Description} \\ 
    \midrule
    student\_catch\_up & Suggests that Jeremy needs to catch up on the academic work. \\
    school\_policy & Mentions the school's policy requiring a doctor's note. \\
    rejects\_policy & Notes that the policy is unfair or discriminatory to the student. \\
    learn\_challenge & Identifies an academic challenge Jeremy faces. \\
    change\_for & Proposes changes to the planned instruction. \\
    more\_some & Argues for giving more academic support to Jeremy or other students who need it. \\ 
    jeremy\_effort & Attributes Jeremy's actions to a lack of effort, focus, or a negative attitude. \\
    jeremy\_mental & Recognizes that something is wrong with Jeremy's mental state. \\ 
    jeremy\_engaged & Recognizes that Jeremy is bored or disengaged with the material. \\ 
    not\_well & Observes or states that Jeremy isn't feeling well. \\ 
    modification & Suggests a modification to the scheduled quiz, such as moving the date, allowing a retake, or not counting the grade. \\ 
    equality & States that Jeremy should take the quiz. \\ 
    feel\_jeremy & Asks Jeremy how he is feeling or states that they hope he is feeling better. \\ 
    teacher\_catch\_up & Offers to meet with Jeremy to help him catch up on missed classwork. \\ 
    \bottomrule
    \end{tabular}
    \caption{Characteristics and their descriptions.}
    \label{tab:characteristics}
\end{table}

Before the data was used for model evaluation, preprocessing was carried out. This included the removal of duplicate responses and the resolution of any encoding errors present in the data. After preprocessing, a total of 1,201 unique responses were obtained. Additionally, to prepare the data for model training and testing, the responses were processed so that each sample in the dataset contained one response paired with one characteristic. This resulted in a total of 4,822 unique response-characteristic pairs. 

\begin{figure}[htbp]
    \centering
    \includegraphics[width=1\textwidth]{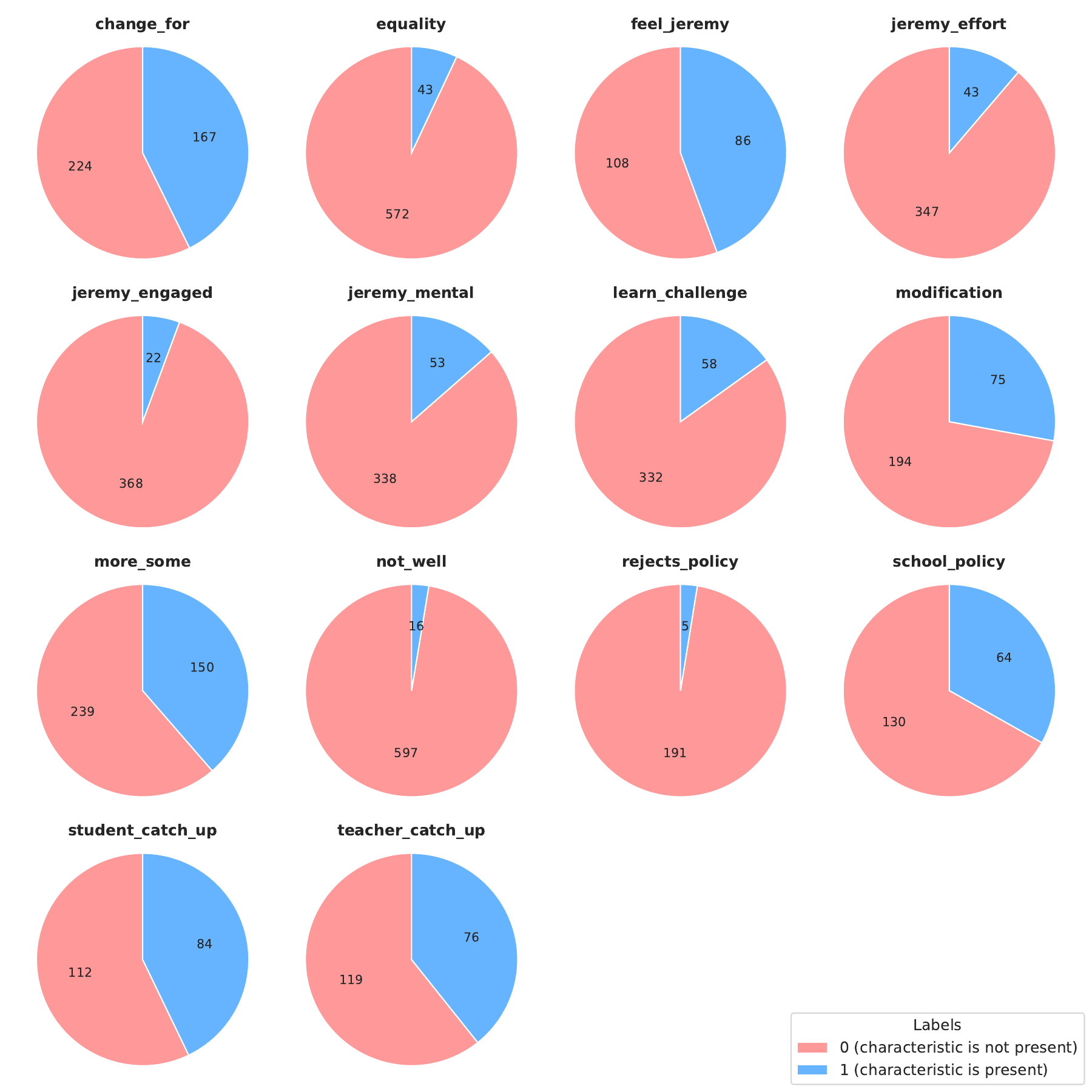}
    \caption{Distribution of label values across characteristics}
    \label{fig:char_distribution}
\end{figure}

In Figure \ref{fig:char_distribution}, the distribution of responses for each characteristic is presented, showing both positive and negative labels. Some characteristics have imbalanced cases, such as ``not\_well", which is heavily skewed with 597 negative labels and just 16 positive labels; or ``rejects\_policy", with 191 negative labels and only 5 positive labels. Conversely, ``change\_for" and ``student\_catch\_up" have more balanced distributions, with a relatively higher number of positive labels. These insights highlight areas where responses from participants varied significantly.

This imbalance indicates that some characteristics are far more frequently observed than others. This could be due to the inherent nature of the characteristics or the specific context of the simulation. Additionally, the skewed distributions might reflect varying levels of clarity or specificity in the characteristics themselves. If characteristics are more ambiguous or open to interpretation, this could lead to inconsistent labeling by the expert evaluators. Such inconsistency in labeling could affect the performance of language models, as the models might struggle to learn from and predict these traits accurately. Therefore, clear and precise definitions for each characteristic are crucial to ensure more reliable and accurate model predictions. These observations about the ambiguity or openness of some characteristics will be discussed in later sections.

\subsection{Dataset splitting}


The dataset was divided into three different splits to facilitate the evaluation of the language models. One split (test set) contains characteristics that are not present in the other two splits. The other two splits (training set and evaluation set) share common characteristics. The process of constructing these subsets involved several steps.

From the full dataset, the samples corresponding to a specific subset of characteristics were extracted to form the first split, referred to as the test set. This test set contains characteristics that do not appear in the remaining splits of the dataset.

The remaining data, which excludes the characteristics in the test set, was then split into two parts: the training set and the evaluation set. This split was done using a 70/30 stratified method based on both the characteristics and their corresponding labels (0 or 1). Stratified splitting means that the data was divided in such a way that both the training and evaluation sets have the same distribution of characteristics. For example, if a specific characteristic is present in 10\% of the data, it will also be present in about 10\% of each of these two sets. In addition, the ratio of zeros (absent) and ones (present) is maintained for each characteristic. This ensures that both sets are representative and balanced, making the scoring both fair and reliable.

The training set, comprising 70\% of this remaining data, was used to train the models. The evaluation set, consisting of the remaining 30\%, was used to assess the performance of the model on characteristics that were present during training.

This approach ensures that the test set contains unique characteristics, while the training and evaluation sets are balanced and representative. This setup allows for a robust assessment of how well the models perform on both familiar and unfamiliar characteristics.

To determine which characteristics to include in the test set, three different strategies were employed. The first strategy, called the ``low-sample experiment", involved selecting three characteristics with a reduced set of samples. This approach was taken to minimize the number of samples in the test set and thus maximize the number available for the training set. By doing so, the model could be trained on a larger dataset, potentially improving its performance.

The second strategy, known as the ``random selection experiment", involved selecting three random characteristics. This was done to gain insights into the importance of different characteristics in the results. By randomly selecting characteristics, the variability in model performance due to the specific characteristics included in the test set could be better understood.

The third strategy, referred to as the ``performance-matched experiment", focused on selecting three characteristics for which the performance of the Llama 3 model in a zero-shot setting was similar in both the test and evaluation sets. This approach was aimed at minimizing the effect of characteristic selection on the results, ensuring that the performance evaluation was not overly influenced by the specific characteristics chosen for the test set.

Table \ref{tab:dataset_experiments} summarizes the distribution of samples across the train, evaluation, and test sets for each of the three strategies. The table includes the total number of samples in each set as well as the specific characteristics selected for the test set for each strategy. This approach allowed for a robust assessment of the capabilities of the models under different conditions.

\begin{table}[htbp]
\centering
\begin{adjustbox}{width=1\textwidth}
\begin{tabular}{l|p{3cm}p{3cm}p{3cm}}
\toprule
& \textbf{Low-sample} & \textbf{Random} & \textbf{Performance-Matched} \\
\midrule
\textbf{Train} & 2,963 & 2,776 & 2,827 \\
\textbf{Evaluation} & 1,271 & 1,190 & 1,212 \\
\textbf{Test} & 588 & 856 & 783 \\
\textbf{Characteristics} & school\_policy rejects\_policy feel\_jeremy & modification teacher\_catch\_up change\_for & jeremy\_mental 
rejects\_policy
feel\_jeremy\\
\bottomrule
\end{tabular}
\end{adjustbox}
\caption{Distribution of samples across train, validation, and test sets for the three different strategies: low-sample, random, and performance-matched}
\label{tab:dataset_experiments}
\end{table}

\subsection{Models and configurations}
Llama 3 \citep{llama3modelcard} and DeBERTaV3 \cite{he2021debertav3} were used in this study. The 8-billion parameter version of Llama 3 was chosen for several reasons. It is one of the latest state-of-the-art open-weight model, making it accessible for research and practical applications. Additionally, Llama 3 is recognized for its high performance and cost-effectiveness. In particular, the instruction-tuned version of Llama 3 was used to enhance its ability to follow and execute specific instructions accurately. Due to GPU memory constraints, the 8-billion parameter version was specifically selected, ensuring effective use within the available computational resources. 

DeBERTaV3 was also selected for its performance as one of the most advanced encoder-only models currently available. This model stands out for its ability to outperform other well-known models such as XLNet \citep{yang2019xlnet}, RoBERTa \citep{liu2019roberta} and ELECTRA \citep{clark2020electra} in various benchmarks. DeBERTaV3 combines the strengths of DeBERTa and ELECTRA using a novel embedding sharing paradigm called GDES. Its ability to handle intricate language understanding tasks with high accuracy made it an ideal choice for our study.


To thoroughly evaluate the performance of these models, five different configurations were tested. Using Llama 3, four configurations were evaluated:

\begin{itemize}
    \item \textbf{Zero-shot configuration}: In this setup, the model is provided only with the instruction, the response, and the characteristic it needs to classify. No additional examples are given.
    \item \textbf{Few-shot configuration}: In addition to the instruction, the response, and the characteristic, the model is provided with five examples of already labeled responses/characteristics from the training set. This helps the model to better understand the task by seeing examples of how similar tasks were labeled \citep{brown2020language}.
    \item \textbf{Fine-tuned zero-shot configuration}: This setup uses the same prompt as the zero-shot configuration, but the model has been previously fine-tuned with the training dataset. The fine-tuning process involves updating the weights of the model weights based on the training data to improve its performance on specific tasks. The following sections will detail the fine-tuning process of the models.
    \item \textbf{Fine-tuned few-shot configuration}: Similar to the few-shot configuration, but using the fine-tuned model. The model receives the instruction, the response, the characteristic, and five examples of labeled responses/characteristics from the training set. The fine-tuning step helps the model perform better by leveraging the specific characteristics of the training data.
\end{itemize}

For the fifth configuration, DeBERTaV3 was used:

\begin{itemize}
    \item \textbf{DeBERTaV3 configuration}: This model is fine-tuned using the training subset. Unlike Llama 3, DeBERTaV3 is an encoder-only model, which fundamentally changes how it can be used. Encoder-only models like DeBERTaV3 are designed for tasks such as classification or sentence encoding, but they cannot generate text or follow instructions in the same way as decoder or encoder-decoder models like Llama 3 \citep{tunstall2022natural} do. This architectural difference means that DeBERTaV3 cannot be used in zero-shot or few-shot configurations that require the model to understand and follow written instructions or examples within the input. Instead, DeBERTaV3 is fine-tuned on the task and then directly provided with the response and the description of the characteristic it needs to classify during inference. This fine-tuning process adapts the pre-trained knowledge of the model to the specific task of characteristic classification, allowing it to make accurate predictions without the need for explicit instructions or examples in the input.
\end{itemize}

Each of these five configurations were tested on both the evaluation and test datasets for each of the three experiments (low-sample, random selection, and performance-matched). This comprehensive testing approach ensures that the performance of the model can be evaluated across different scenarios and datasets, providing a robust assessment of their capabilities. Using these configurations, the research questions can be effectively addressed, allowing the determination of the most effective LLM configuration for both predefined and open-ended identification of characteristics within the TM platform. The prompts used for few-shot and zero-shot configurations can be found in the appendix (see \ref{appendix:prompt_design}).

In Figure \ref{fig:model-configurations}, these configurations and the experimental setup are graphically illustrated, showing the training, evaluation on the same labels, and evaluation on new labels.

To provide a more comprehensive evaluation of current language model capabilities, we extended our analysis to include two additional state-of-the-art decoder-only models: Phi-4-mini \citep{microsoft2025phi4minitechnicalreportcompact} and Qwen 3 8B \citep{qwen3}. Both of these models represents the very latest advancements in language model architecture and capabilities available at the time of our research.

Phi-4-mini \citep{microsoft2025phi4minitechnicalreportcompact} is a compact model with 3.8 billion parameters that has demonstrated strong reasoning and instruction-following capabilities despite its relatively small size. This model was selected to evaluate whether smaller, more efficient models could perform comparably to larger ones on educational characteristic identification tasks, which would have implications for deployment in resource-constrained educational settings.

Qwen 3 8B \citep{qwen3}, with 8 billion parameters (comparable to our primary Llama 3 model), represents an alternative model architecture with its own unique training methodology. Including this model allows us to assess whether our findings generalize across different model families with similar parameter counts, or if there are architecture-specific advantages in characteristic identification tasks.

Both these additional models were evaluated using the zero-shot and few-shot configurations following the same methodology applied to Llama 3, but without fine-tuning, to assess their out-of-the-box capabilities for educational characteristic identification. This approach helps determine whether recent advancements in base model pre-training and instruction tuning can reduce the need for task-specific fine-tuning in educational applications.

\begin{figure}[htbp]
    \centering
    \includegraphics[width=0.90\textwidth]{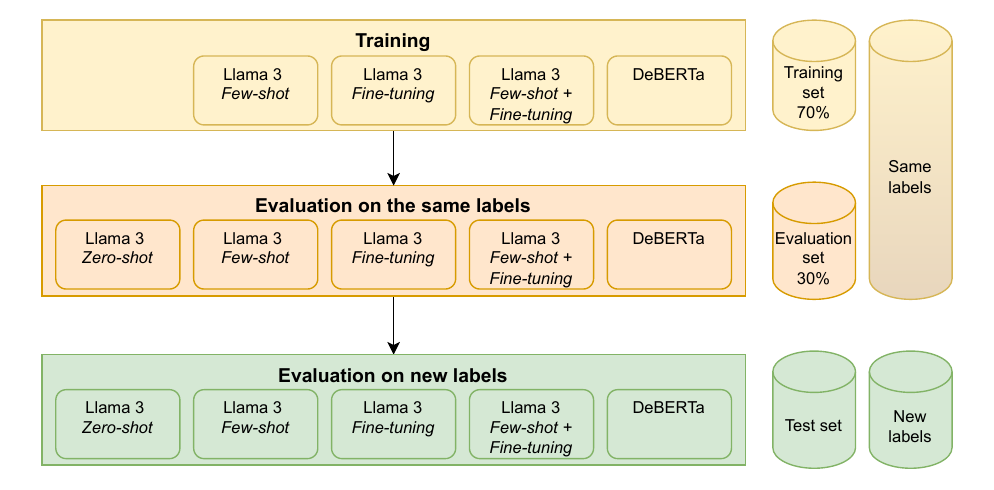}
    \caption{Graphical illustration of model configurations and experimental setup for training, evaluation on the same labels, and evaluation on new labels}
    \label{fig:model-configurations}
\end{figure}

\subsection{Inference and fine-tuning methods}

Specific inference and fine-tuning methods were used to evaluate the models. These methods were designed to ensure accurate and efficient model performance for a binary classification task.

For the inference of Llama 3, the vLLM tool was used \citep{kwon2023efficient} with the greedy-search decoding method. To determine if the characteristic was present in the response, the model was prompted to generate a ``1" if the characteristic was present and a ``0" if it was not. A guided choice method was employed to ensure the model generated either a ``0" or a ``1". This involved focusing only on the probabilities of the tokens corresponding to ``0" and ``1", normalizing these probabilities, and then selecting the token with the highest normalized probability. This approach ensured that the model's output was restricted to the desired binary classification. This deterministic inference process for Llama 3, combining greedy search and a guided choice over the specific target tokens (``0'' or ``1''), ensures reproducible classification outputs and makes temperature settings (typically used to control randomness in broader generative tasks) irrelevant for this specific binary classification.

For the inference of DeBERTaV3, the Transformers library was used \citep{wolf-etal-2020-transformers}. Since the model was already fine-tuned and included a binary classification head, the inference process was straightforward. The model was directly provided with the response and the description of the characteristic, and it outputted the classification without any additional steps. This approach is consistent with the architecture of DeBERTaV3 as an encoder-only model, which, as previously mentioned, is not suitable for zero-shot or few-shot configurations and it is instead used in its fine-tuned state for direct classification tasks.

The fine-tuning of Llama 3 involved a supervised fine-tuning (SFT) approach. The same prompt used for the zero-shot configuration was applied. During fine-tuning, all tokens corresponding to the prompt were masked to ensure that weight updates were based solely on the prediction of the label (0 or 1). The Trainer API from the Transformers library was used too. Additionally, the QLoRA technique \citep{dettmers2023qlora} was employed to reduce the computational resources and time required for training. QLoRA is a combination of two techniques: Low-Rank Adaptation (LoRA) \citep{hu2021lora} and 4-bit quantization \citep{dettmers2023case}. This technique helps optimize the model efficiently by reducing the model size and accelerating the training process.

For the fine-tuning of DeBERTaV3, the input included both the response and the description of the characteristic. A binary classification head, initialized randomly, was added to the pre-trained model. Full-parameter fine-tuning was then performed, meaning the entire model, including the new classification head, was trained. The Trainer API from the Transformers library was used for this training process \citep{wolf-etal-2020-transformers}.

\subsection{Evaluation procedure}

The evaluation of the models was conducted using four key metrics: precision, recall, F1 score, and balanced accuracy. These metrics provide a comprehensive assessment of the models' performance in classifying the characteristics accurately.

\textbf{Precision} measures the proportion of true positive predictions among all positive predictions (true positives plus false positives) made by the model. It indicates how many of the positive identifications were actually correct. High precision is crucial when the cost of false positives is high.

\textbf{Recall} (or sensitivity) measures the proportion of true positive predictions among all actual positives. It reflects the model's ability to identify all relevant instances. High recall is important when the cost of false negatives is high.

\textbf{F1 Score} is the harmonic mean of precision and recall. It provides a single metric that balances both precision and recall, making it useful when a balance between the two is desired. The F1 score is particularly useful when the class distribution is imbalanced.

\textbf{Balanced Accuracy} is the average of recall obtained on each class. It adjusts for imbalanced class distributions by considering both the sensitivity and the specificity (true negative rate) of the model. This metric is particularly useful in scenarios where class imbalance is present. Unlike the F1 score, which balances precision and recall, balanced accuracy takes into account both the positive and negative classes, making it more suitable for assessing performance across all classes in an imbalanced dataset.

These metrics were chosen to provide a comprehensive evaluation of the models. Precision and recall provide insight into the types of errors the models make, while the F1 score provides a balanced view of the overall performance. Balanced accuracy is included to account for any class imbalance in the datasets, ensuring that the performance of the models is evaluated fairly across all classes.

\section{Results}
\label{sec:results}

In this section, the results of the experiments are presented. Tables \ref{tab:performance_metrics_low_sample_test} to \ref{tab:performance_metrics_performance_matched_evaluation} show the performance metrics, including balanced accuracy (BAC), F1 score, precision, and recall, for each of the three experiments: low-sample, random, and performance-matched. Each experiment was evaluated using five configurations (Llama 3 zero-shot, Llama 3 few-shot, Llama 3 fine-tuned zero-shot, Llama 3 fine-tuned few-shot, and DeBERTaV3) on both the evaluation and test datasets.

\begin{table}[h]\centering
\begin{tabular}{l|cccc}
\toprule
Experiment & BAC & F1 & Precision & Recall \\
\midrule
Llama 3 zero-shot & 0.796 & 0.740 & \textbf{0.969} & 0.599 \\
Llama 3 few-shot & 0.913 & 0.884 & 0.918 & 0.854 \\
Llama 3 fine-tuned zero-shot & 0.895 & 0.870 & 0.941 & 0.809 \\
Llama 3 fine-tuned few-shot & \textbf{0.926} & \textbf{0.899} & 0.920 & \textbf{0.879} \\
DeBERTaV3 & 0.762 & 0.656 & 0.683 & 0.631 \\
\bottomrule
\end{tabular}
\caption{Performance metrics for low-sample experiment (test dataset)}
\label{tab:performance_metrics_low_sample_test}
\end{table}

\begin{table}[h]\centering
\begin{tabular}{l|cccc}
\toprule
Experiment & BAC & F1 & Precision & Recall \\
\midrule
Llama 3 zero-shot & 0.732 & 0.526 & 0.437 & 0.660 \\
Llama 3 few-shot & 0.742 & 0.530 & 0.426 & 0.702 \\
Llama 3 fine-tuned zero-shot & \textbf{0.867} & \textbf{0.744} & \textbf{0.678} & 0.824 \\
Llama 3 fine-tuned few-shot & 0.859 & 0.676 & 0.547 & \textbf{0.887} \\
DeBERTaV3 & 0.819 & 0.684 & 0.642 & 0.731 \\
\bottomrule
\end{tabular}
\caption{Performance metrics for low-sample experiment (evaluation dataset)}
\label{tab:performance_metrics_low_sample_evaluation}
\end{table}

\begin{table}[h]\centering
\begin{tabular}{l|cccc}
\toprule
Experiment & BAC & F1 & Precision & Recall \\
\midrule
Llama 3 zero-shot & 0.691 & 0.603 & 0.647 & 0.564 \\
Llama 3 few-shot & 0.731 & 0.671 & 0.606 & \textbf{0.752} \\
Llama 3 fine-tuned zero-shot & 0.621 & 0.428 & \textbf{0.76} & 0.298 \\
llama 3 fine-tuned few-shot & \textbf{0.764} & \textbf{0.703} & 0.704 & 0.702 \\
DeBERTaV3 & 0.597 & 0.398 & 0.643 & 0.288 \\
\bottomrule
\end{tabular}
\caption{Performance metrics for random experiment (test dataset)}
\label{tab:performance_metrics_random_test}
\end{table}

\begin{table}[h]\centering
\begin{tabular}{l|cccc}
\toprule
Experiment & BAC & F1 & Precision & Recall \\
\midrule
Llama 3 zero-shot & 0.752 & 0.519 & 0.419 & 0.683 \\
Llama 3 few-shot & 0.731 & 0.480 & 0.372 & 0.677 \\
Llama 3 fine-tuned zero-shot & 0.773 & 0.665 & \textbf{0.784} & 0.577 \\
Llama 3 fine-tuned few-shot & 0.828 & 0.655 & 0.574 & \textbf{0.762} \\
DeBERTaV3 & \textbf{0.834} & \textbf{0.723} & 0.727 & 0.720 \\
\bottomrule
\end{tabular}
\caption{Performance metrics for random experiment (evaluation dataset)}
\label{tab:performance_metrics_random_evaluation}
\end{table}

\begin{table}[h]\centering
\begin{tabular}{l|cccc}
\toprule
Experiment & BAC & F1 & Precision & Recall \\
\midrule
Llama 3 zero-shot & 0.742 & 0.537 & 0.445 & 0.676 \\
Llama 3 few-shot & \textbf{0.882} & 0.746 & 0.653 & \textbf{0.869} \\
Llama 3 fine-tuned zero-shot & 0.809 & 0.729 & 0.832 & 0.648 \\
Llama 3 fine-tuned few-shot & 0.838 & \textbf{0.77} & \textbf{0.85} & 0.703 \\
DeBERTaV3 & 0.669 & 0.471 & 0.535 & 0.421 \\
\bottomrule
\end{tabular}
\caption{Performance metrics for performance-matched experiment (test dataset)}
\label{tab:performance_metrics_performance_matched_test}
\end{table}

\begin{table}[h]\centering
\begin{tabular}{l|cccc}
\toprule
Experiment & BAC & F1 & Precision & Recall \\
\midrule
Llama 3 zero-shot & 0.719 & 0.526 & 0.456 & 0.622 \\
Llama 3 few-shot & 0.729 & 0.517 & 0.401 & 0.726 \\
Llama 3 fine-tuned zero-shot & 0.833 & 0.727 & 0.715 & 0.739 \\
Llama 3 fine-tuned few-shot & \textbf{0.853} & 0.675 & 0.543 & \textbf{0.892} \\
DeBERTaV3 & 0.834 & \textbf{0.744} & \textbf{0.767} & 0.722 \\
\bottomrule
\end{tabular}
\caption{Performance metrics for performance-matched experiment (evaluation dataset)}
\label{tab:performance_metrics_performance_matched_evaluation}
\end{table}

\paragraph{Low-Sample experiment}

In the low-sample experiment (Table \ref{tab:performance_metrics_low_sample_test} and Table \ref{tab:performance_metrics_low_sample_evaluation}), it is observed that the Llama 3 fine-tuned few-shot configuration consistently performed best on the test dataset with a BAC of 0.926, an F1 score of 0.899, and the highest recall of 0.879. This suggests that fine-tuning with few-shot examples significantly improves the capability of the model to generalize to unseen characteristics.

On the other hand, the Llama 3 fine-tuned zero-shot configuration showed the highest performance in the evaluation set with a BAC of 0.867 and an F1 score of 0.744, indicating that the model retains strong performance even without additional few-shot examples when tested on seen characteristics. DeBERTaV3, while performing reasonably well, fell behind the Llama 3 configurations, particularly in the test dataset where it achieved a BAC of 0.762 and an F1 score of 0.656.

\paragraph{Random experiment}

The random experiment (Table \ref{tab:performance_metrics_random_test} and Table \ref{tab:performance_metrics_random_evaluation}) highlighted some variability in model performance. On the test dataset, Llama 3 fine-tuned few-shot achieved the best results with a BAC of 0.764 and an F1 score of 0.703, showing its robustness across randomly selected characteristics. Notably, the Llama 3 fine-tuned zero-shot configuration performed poorly in this scenario with a BAC of 0.621 and an F1 score of 0.428, indicating challenges in generalizing without few-shot examples. Interestingly, Llama 3 fine-tuned zero-shot achieved the highest precision (0.76), while Llama 3 few-shot had the highest recall (0.752). This suggests that the zero-shot configuration was more conservative in its predictions, making fewer but more accurate positive classifications. In contrast, the few-shot configuration was more sensitive, correctly identifying a higher proportion of positive cases but at the cost of more false positives.

In the evaluation set, DeBERTaV3 exhibited the highest BAC (0.834) and F1 score (0.723). Llama 3 fine-tuned few-shot also demonstrated strong performance with a BAC of 0.828 and an F1 score of 0.655. Noteworthy, Llama 3 fine-tuned few-shot achieved the highest recall (0.762), indicating its effectiveness in identifying a larger proportion of positive cases. On the other hand, Llama 3 fine-tuned zero-shot showed the best precision (0.784), suggesting more accurate positive predictions but potentially at the cost of missing some positive cases.

\paragraph{Performance-matched experiment}

For the performance-matched experiment (Table \ref{tab:performance_metrics_performance_matched_test} and Table \ref{tab:performance_metrics_performance_matched_evaluation}), the Llama 3 few-shot configuration scored highest on the test dataset with a BAC of 0.882 and a recall of 0.869, highlighting its ability to maintain high recall across conditions. The fine-tuned few-shot configuration also performed well with a BAC of 0.838 and the highest F1 score of 0.770. In addition, this configuration also achieved the highest precision of 0.85, indicating its balanced performance across different metrics and its ability to make accurate positive predictions.

In the evaluation set, DeBERTaV3 achieved the highest F1 score of 0.744 and Precision of 0.767, suggesting its strong performance with characteristics seen during training. However, Llama 3 fine-tuned few-shot achieved the highest BAC of 0.853 and Recall of 0.892, indicating its effectiveness in handling seen characteristics using fine-tuning and few-shot examples.

\paragraph{Comprehensive model comparison}

\begin{figure}[htbp]
    \centering
    \includegraphics[width=0.9\textwidth]{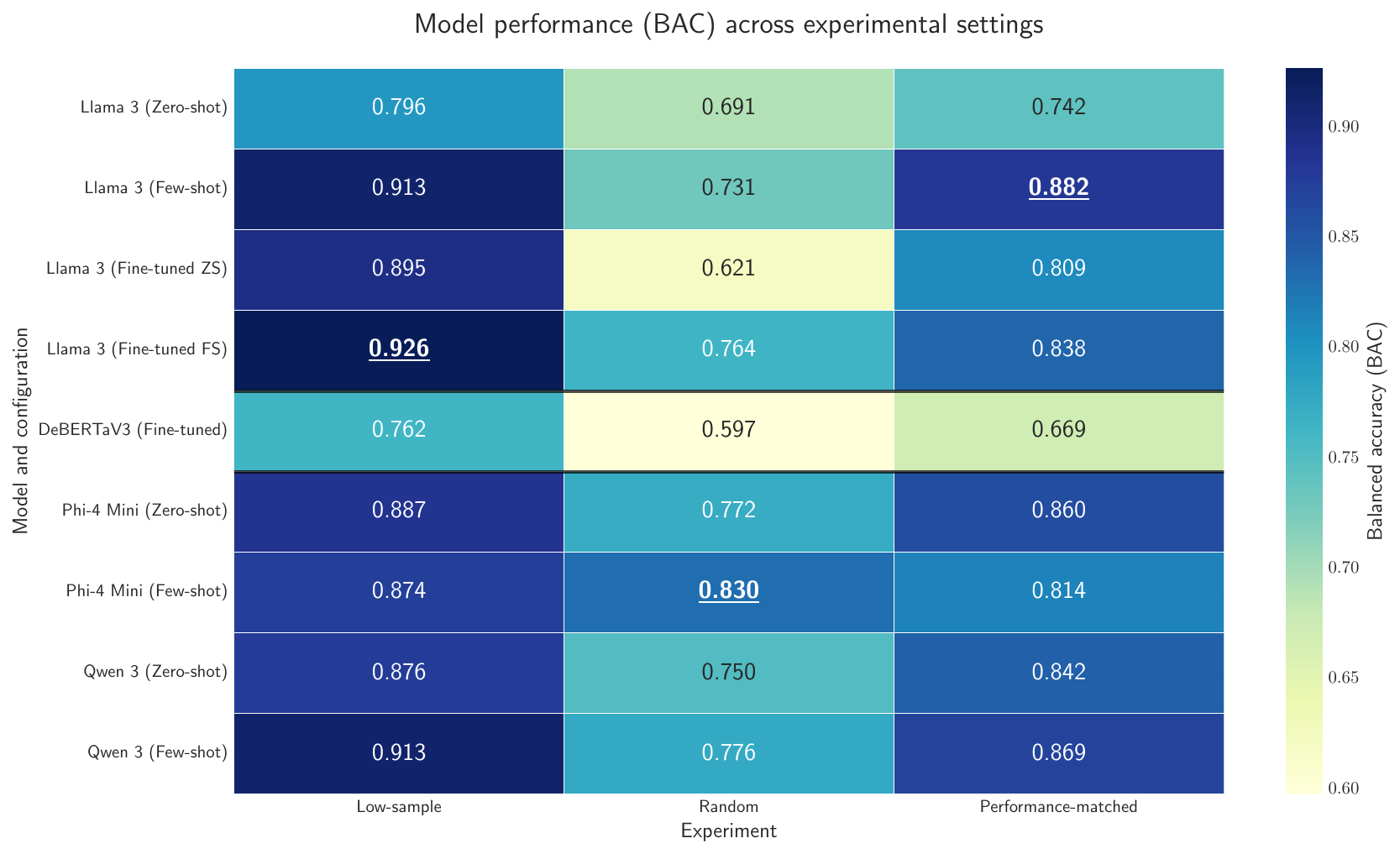}
    \caption{Heatmap of Balanced Accuracy BAC scores across all model configurations and experimental settings. The highest values for each experimental setting are underlined and displayed in bold.}
    \label{fig:bac_heatmap}
\end{figure}

To provide a clearer comparison across all evaluated models and experimental settings, Figure \ref{fig:bac_heatmap} presents a heatmap visualization of the BAC scores for all model configurations across the three experimental scenarios, focusing exclusively on performance with the test datasets. This visualization allows for immediate identification of performance patterns and highlights the best-performing configuration for each experiment with underlined and bold text.

\begin{figure}[htbp]
    \centering
    \includegraphics[width=0.9\textwidth]{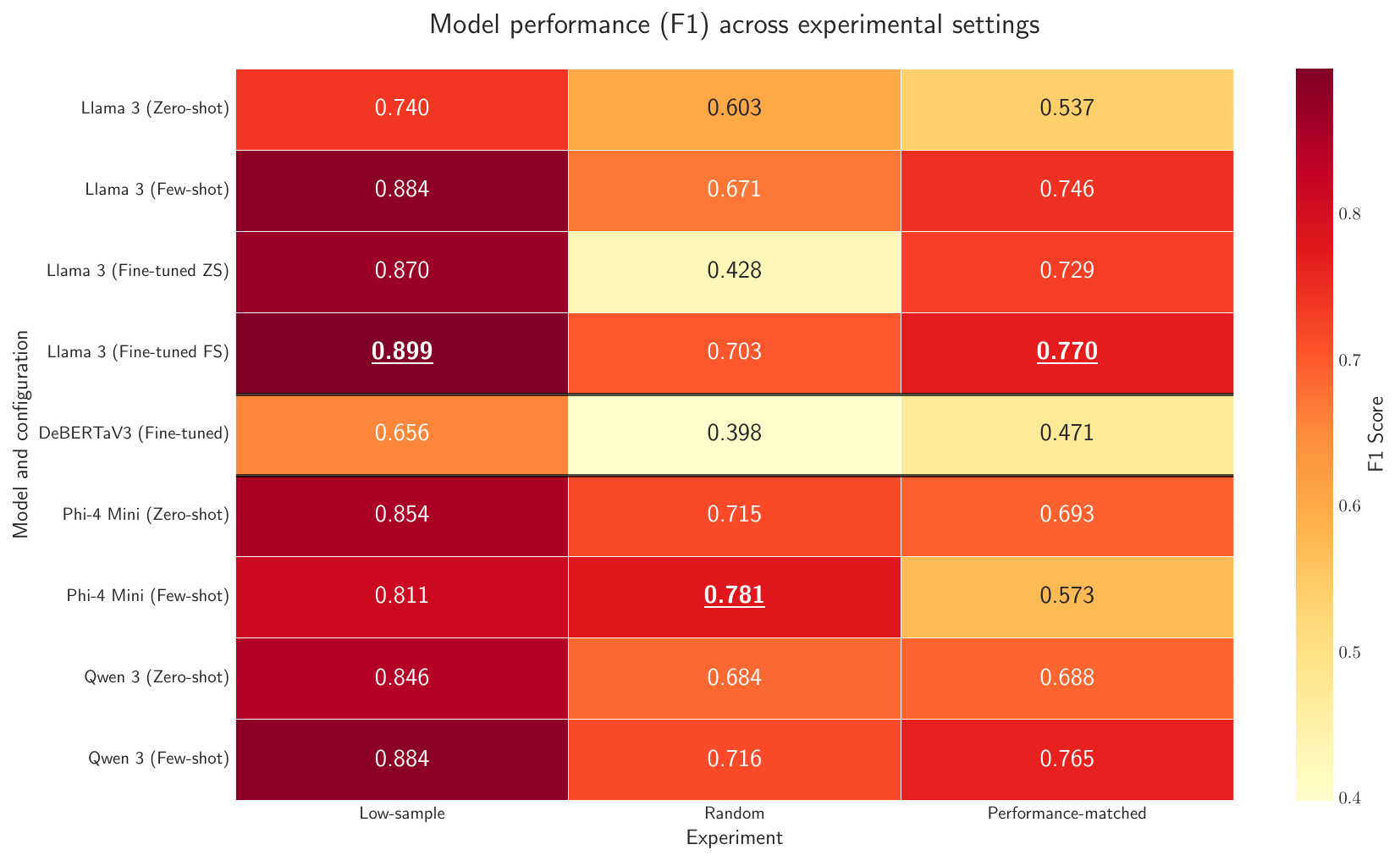}
    \caption{Heatmap of F1 scores across all model configurations and experimental settings. The highest values for each experimental setting are underlined and displayed in bold.}
    \label{fig:f1_heatmap}
\end{figure}

Similarly, Figure \ref{fig:f1_heatmap} presents the F1 scores for all configurations, providing insight into the precision-recall balance achieved by each model.

As shown in the heatmaps, the newly evaluated models Phi-4-mini and Qwen 3 demonstrate competitive performance compared to the original models. This is particularly evident in their zero-shot capabilities, where they consistently outperform Llama 3's zero-shot performance. Notably, Phi-4-mini achieves the highest BAC score (0.887) in the low-sample experiment using only zero-shot configuration, outperforming all non-fine-tuned approaches and even rivaling some fine-tuned configurations. This suggests significant advancements in the inherent capabilities of recent language models for educational characteristic identification without task-specific optimization. For a comprehensive view of all performance metrics across both test and evaluation datasets for all models and configurations, we provide a detailed table in Appendix Table \ref{tab:appendix_detailed_results}.

\section{Discussion} 
\label{sec:discussion}

The results indicate that the Llama 3 fine-tuned few-shot configuration generally provides the best performance across different scenarios, especially for unseen characteristics in the test datasets. This configuration showed consistently high BAC and F1 scores, highlighting the benefits of fine-tuning combined with few-shot learning to improve the generalization ability of the model.

DeBERTaV3, while effective, showed a more significant drop in performance on the test datasets compared to the evaluation sets, indicating it might be less robust to predict completely new characteristics. However, it still maintained competitive performance on the evaluation sets, especially in the performance-matched and random experiments.

Llama 3 zero-shot and few-shot configurations performed variably, with few-shot generally outperforming zero-shot, suggesting that providing additional examples helps the model understand and classify the characteristics more accurately. Overall, these findings suggest that fine-tuning and few-shot learning are critical for improving the generalization capabilities of language models. The use of these techniques allows models to adapt better to new, unseen characteristics, making them more reliable and effective in practical applications.

The comprehensive heatmap visualizations (Figures \ref{fig:bac_heatmap} and \ref{fig:f1_heatmap}) provide several valuable insights that deepen our understanding of model performance across experimental settings. First, examining the patterns within the BAC heatmap reveals a clear stratification of performance across model families and configurations. The fine-tuned models generally occupy the upper performance tiers, with Llama 3 fine-tuned few-shot configuration achieving the highest BAC (0.926) in the low-sample experiment. However, the distribution of performance is not uniform across all experiments, suggesting that different approaches may be optimal depending on the specific characteristics being identified.

One particularly notable finding is the great zero-shot performance of Phi-4-mini, especially in the low-sample experiment where it achieved a BAC of 0.887 without any fine-tuning or in-context examples. This is remarkable considering its relatively small size (3.8B parameters) compared to Llama 3 (8B parameters). This suggests that recent architectural innovations and training methodologies can significantly improve parameter efficiency for educational characteristic identification tasks. The ability of smaller models to perform comparably to larger ones has important implications for the practical deployment of LLMs in educational settings where computational resources may be limited.

The heatmaps also reveal interesting patterns across experimental types. All models, regardless of architecture or configuration, performed better in the low-sample experiment compared to the random experiment. This consistency across model families reinforces our answer to RQ1: the effectiveness of characteristic identification depends significantly on the nature of the characteristic itself. Some characteristics appear inherently more identifiable than others, regardless of the model or approach used.

Examining the F1 score heatmap alongside the BAC heatmap provides additional nuance to our understanding. While BAC measures overall classification accuracy accounting for class imbalance, F1 scores reflect the precision-recall trade-off. For example, in the low-sample experiment, Phi-4-mini zero-shot achieves an F1 score of 0.854, nearly matching its BAC of 0.887, indicating a good balance between precision and recall. In contrast, in the performance-matched experiment, Phi-4-mini few-shot shows a much larger gap between its BAC (0.814) and F1 (0.573), suggesting it may be achieving high balanced accuracy at the cost of precision.

The comparative performance between few-shot and zero-shot configurations across models is also revealing. While few-shot prompting generally improves performance for Llama 3 and Qwen 3, interestingly, Phi-4-mini sometimes performs better with zero-shot prompting. In the low-sample experiment, Phi-4-mini's zero-shot BAC (0.887) exceeds its few-shot BAC (0.874), and a similar pattern appears in the performance-matched experiment. This suggests that providing examples may occasionally constrain the model's reasoning or introduce biases that reduce performance, particularly for models with strong zero-shot capabilities.

Lastly, when comparing non-fine-tuned models, we observe that both Phi-4-mini and Qwen 3 consistently outperform the baseline Llama 3 in zero-shot and few-shot configurations. This highlights the rapid pace of advancement in language model capabilities, where newer models demonstrate superior out-of-the-box performance for specialized tasks like educational characteristic identification. This trend has important implications for our research questions, particularly RQ3, suggesting that as base models continue to improve, the necessity for fine-tuning may diminish for certain applications, potentially making these technologies more accessible for educational contexts with limited resources for model customization.

The performance variations observed across different models (as detailed in Appendix Table \ref{tab:appendix_detailed_results}) likely arise from differences in pre-training datasets, architectural innovations, and post-training procedures such as instruction tuning and alignment. Phi-4-mini's \citep{microsoft2025phi4minitechnicalreportcompact} superior zero-shot performance despite fewer parameters suggests more efficient architecture and potentially higher-quality training data. However, without access to proprietary training details, these remain informed hypotheses.

A notable comparison can be made with the work of \citet{LittenbergTobias2022}, where researchers used the weighted F1 score to evaluate their binary classification models for detecting specific educational characteristics in teacher responses. Their study used the same dataset source, meaning that many of the characteristics in our study overlap with theirs. In their work, they compared the performance of GPT-3 with traditional machine learning algorithms and found that the latter performed better. The task in both their study and our current research is the same: determining whether a specific characteristic is present in teacher candidates' responses. While the weighted F1 score provides a holistic view of the performance of the model across both classes, it may not be the most effective for imbalanced datasets where the primary interest is the positive class. The weighted average can obscure the capacity of the model to correctly identify the less frequent but critical educational characteristics.

In contrast, the binary F1 score (the one used in this study) offers a clearer evaluation for such scenarios. This metric focuses specifically on the positive class, balancing precision and recall to give a more accurate measure of how well the model identifies the presence of the characteristics. Emphasizing the positive class ensures that both false positives and false negatives are adequately considered, providing a more meaningful assessment of the effectiveness of the model in real-world educational settings.

To compare our approach with that of \citet{LittenbergTobias2022}, the results were computed using the weighted F1 score and grouped by the same characteristics. Table \ref{tab:comparison_table} shows the results from the compared study alongside those of the fine-tuned Llama 3 model for each of the three experiments in the zero-shot and few-shot configurations. For this comparison, we focused on the fine-tuned versions of Llama 3, as these configurations demonstrated strong performance in our experiments. The Llama 3 model was fine-tuned using the training dataset, and the results shown are from the evaluation dataset. The ``-" symbol indicates characteristics that are not present in the evaluation set for that specific experiment. This is because certain characteristics were reserved for the test subset, to evaluate its performance on unseen characteristics.

\begin{table}[h]
\centering
\begin{adjustbox}{width=1\textwidth}
\begin{tabular}{l|cccccccccc}
\toprule
Approach & 1 & 2 & 3 & 4 & 5 & 6 & 7 & 8 & 9 & Average \\
\midrule
Human Baseline \citep{LittenbergTobias2022} & 0.93 & 0.84 & 0.78 & 0.92 & 0.78 & 0.79 & 0.77 & 0.88 & 0.86 & 0.84 \\
Best Model \citep{LittenbergTobias2022} & 0.94 & 0.72 & 0.69 & \textbf{0.95} & 0.70 & 0.69 & 0.75 & 0.84 & 0.86 & 0.79 \\
Best Model (Ours) & \textbf{0.97} & \textbf{0.90} & \textbf{0.86} & 0.93 & \textbf{0.88} & \textbf{0.89} & \textbf{0.78} & \textbf{0.92} & \textbf{0.88} & \textbf{0.89} \\
Low-sample - few shot & - & 0.87 & 0.74 & - & 0.80 & 0.71 & 0.66 & 0.92 & 0.83 & 0.79 \\
Low-sample - zero-shot & - & 0.90 & 0.81 & - & 0.88 & 0.89 & 0.76 & 0.90 & 0.86 & 0.86 \\
Random - few shot & 0.93 & - & 0.81 & 0.93 & 0.87 & - & 0.60 & 0.85 & 0.83 & 0.83 \\
Random - zero shot & 0.97 & - & 0.86 & 0.93 & 0.81 & - & 0.78 & 0.89 & 0.84 & 0.87 \\
Performance-matched - few shot & - & 0.85 & 0.81 & 0.92 & 0.77 & 0.62 & 0.52 & - & 0.88 & 0.77 \\
Performance-matched - zero shot & - & 0.90 & 0.81 & 0.93 & 0.88 & 0.77 & 0.76 & - & 0.88 & 0.85 \\
\bottomrule
\end{tabular}
\end{adjustbox}
\caption{Comparison of weighted F1 scores for characteristic detection between the human baseline, best results from \citet{LittenbergTobias2022}, and fine-tuned Llama 3 model. Column labels are 1-feel\_jeremy, 2-teach\_catch\_up, 3-student\_catch\_up, 4-school\_policy, 5-learn\_challenge, 6-change\_for, 7-more\_some, 8-jeremy\_mental, 9-jeremy\_effort. See Table \ref{tab:characteristics} for the description of these characteristics. The best performing model for each characteristic is indicated in bold.}
\label{tab:comparison_table}
\end{table}

The row labeled as ``Best Model (Ours)" refers to the highest F1 scores achieved by the fine-tuned Llama 3 model across any configuration and experiment. Similarly, the ``Best Model" row from \citet{LittenbergTobias2022} represents the best results obtained from any configuration tested in their study. The values highlighted in bold represent the highest F1 scores for each characteristic across all approaches, indicating which configurations and models performed best in detecting those specific educational characteristics.

The fine-tuned Llama 3 model outperforms traditional machine learning methods and even the human baseline in several instances. Notably, Llama 3 achieves higher F1 scores across most characteristics, demonstrating its effectiveness. We also applied a statistical test to assess whether the fine-tuned Llama 3 significantly outperformed the human baseline. Since the assumption of normality of the differences was satisfied, we conducted a paired t-test. The test revealed a significant difference between the performances of the human baseline and Llama 3, with a p-value of 0.0010 and a t-value of 4.7696. 

These findings highlight the potential of using advanced language models like Llama 3, particularly when fine-tuned with domain-specific data, to improve the identification of educational characteristics. This approach not only has the potential to enhance the accuracy of feedback provided to teachers but also underscores the importance of model selection and fine-tuning in developing effective NLP tools for educational settings.

To explore the role of explicit reasoning in educational contexts, we performed additional structured reasoning experiments using Qwen 3. This model features a hybrid thinking mode that allows the model to generate explicit reasoning chains before producing final answers. For this experiment, we activated this thinking mode, enabling the model to generate step-by-step reasoning traces enclosed within special tokens, allowing us to observe its decision-making process. The model showed ability to infer implicit characteristics from teacher responses (e.g., recognizing that partial quiz exemption constitutes a grade modification) but sometimes exhibited over-interpretation, prioritizing literal definitions over practical educational intent. Complete reasoning traces are provided in \ref{appendix:cot_examples}.

Quantitatively, structured reasoning showed mixed results (detailed in Appendix Table \ref{tab:appendix_detailed_results}): while it improved performance in some configurations (e.g., +0.043 BAC improvement in the random few-shot test scenario), it also showed degradation in others (-0.070 BAC in performance-matched zero-shot test), with an overall average change of -0.012 BAC across all configurations. This suggests that explicit reasoning can enhance model interpretability without necessarily improving classification performance, indicating that the value of structured reasoning may lie more in providing educational insights and explanations rather than purely optimizing accuracy metrics.

\section{Conclusions and future work}
\label{sec:conclusions}

DS usually includes open-ended questions enabling the teacher candidates to express their own thoughts. In these simulations, teacher educators need to identify specific behaviors or characteristics in the teacher candidates' responses to provide adequate evaluation and avoid the repetition of mistakes. To address this issue, we have evaluated different configurations of LLMs to identify characteristics in the responses of DS, using a dataset from DS for teacher education. With this dataset, we evaluated the performance of DeBERTaV3 and Llama 3, combined with zero-shot, few-shot, and fine-tuning. We conducted three experiments evaluating the models' performance in three different dataset splits. In these experiments, we discovered a significant variation in the LLMs' performance depending on the characteristic to identify. Experiments with Phi-4-mini and Qwen-3 confirmed that the observed performance variations extended to other LLMs beyond Llama 3 and DeBERTaV3. Additionally, structured reasoning experiments with Qwen 3 revealed that while explicit reasoning traces provide valuable insights into model decision-making processes, they do not consistently improve classification performance, with mixed results across different experimental configurations. Besides, we noted that DeBERTaV3 obtained a balanced accuracy of 0.8 in identifying labels seen during its training but on unseen responses. For new labels, this model decreased its performance, even achieving an accuracy of 0.597 in one of the experiments. In contrast, Llama 3 performed better than DeBERTaV3 in predicting new characteristics and showing more stable performance. In most of the experiments, few-shot and fine-tuning improved the performance of Llama 3 for the identification of both seen labels and new labels. Consequently, we recommend using Llama 3 for the DS where teacher educators need to introduce new characteristics, since they could change depending on the simulation or the educational objectives.

This work could be extended by testing LLMs in other simulation environments. An interesting gap for research is to evaluate the generalization ability of few-shot and fine-tuned LLMs to other simulations designed for other professionals and characteristics. Furthermore, future work could include exploring why some characteristics are more difficult to identify. This work could provide recommendations for educators on defining these characteristics. 

The structured reasoning capabilities demonstrated by Qwen 3 could enable automated feedback generation systems that leverage explicit reasoning traces to provide personalized pedagogical recommendations to teacher candidates. Another promising direction would be developing real-time reasoning explanations that could be integrated directly into simulation platforms, transforming the model from a simple classifier into an interactive pedagogical tool that supports reflection and learning through detailed explanations of identified teaching behaviors.

Additionally, future researchers could evaluate the adoption and perception of educators regarding the integration of LLMs in DS to provide the required evaluations that these environments need for a comprehensive education.

\newpage

\bibliographystyle{elsarticle-num-names} 
\bibliography{cas-refs}

@article{timotheou2023impacts,
  title={Impacts of digital technologies on education and factors influencing schools' digital capacity and transformation: A literature review},
  author={Timotheou, Stella and Miliou, Ourania and Dimitriadis, Yiannis and Sobrino, Sara Villagr{\'a} and Giannoutsou, Nikoleta and Cachia, Romina and Mon{\'e}s, Alejandra Mart{\'\i}nez and Ioannou, Andri},
  journal={Education and information technologies},
  volume={28},
  number={6},
  pages={6695--6726},
  year={2023},
  publisher={Springer},
  doi={10.1007/s10639-022-11431-8} 
}

@misc{qwen3,
    title  = {Qwen3},
    url    = {https://qwenlm.github.io/blog/qwen3/},
    author = {Qwen Team},
    month  = {April},
    year   = {2025}
}

@misc{microsoft2025phi4minitechnicalreportcompact,
      title={Phi-4-Mini Technical Report: Compact yet Powerful Multimodal Language Models via Mixture-of-LoRAs}, 
      author={Abdelrahman Abouelenin and Atabak Ashfaq and Adam Atkinson and Hany Awadalla and Nguyen Bach and Jianmin Bao and Alon Benhaim and Martin Cai and Vishrav Chaudhary and Congcong Chen and Dong Chen and Dongdong Chen and Junkun Chen and Weizhu Chen and Yen-Chun Chen and Yi-ling Chen and Qi Dai and Xiyang Dai and Ruchao Fan and Mei Gao and Min Gao and Amit Garg and Abhishek Goswami and Junheng Hao and Amr Hendy and Yuxuan Hu and Xin Jin and Mahmoud Khademi and Dongwoo Kim and Young Jin Kim and Gina Lee and Jinyu Li and Yunsheng Li and Chen Liang and Xihui Lin and Zeqi Lin and Mengchen Liu and Yang Liu and Gilsinia Lopez and Chong Luo and Piyush Madan and Vadim Mazalov and Arindam Mitra and Ali Mousavi and Anh Nguyen and Jing Pan and Daniel Perez-Becker and Jacob Platin and Thomas Portet and Kai Qiu and Bo Ren and Liliang Ren and Sambuddha Roy and Ning Shang and Yelong Shen and Saksham Singhal and Subhojit Som and Xia Song and Tetyana Sych and Praneetha Vaddamanu and Shuohang Wang and Yiming Wang and Zhenghao Wang and Haibin Wu and Haoran Xu and Weijian Xu and Yifan Yang and Ziyi Yang and Donghan Yu and Ishmam Zabir and Jianwen Zhang and Li Lyna Zhang and Yunan Zhang and Xiren Zhou},
      year={2025},
      eprint={2503.01743},
      archivePrefix={arXiv},
      primaryClass={cs.CL},
      url={https://arxiv.org/abs/2503.01743}, 
}

@book{tunstall2022natural,
  title={Natural language processing with transformers},
  author={Tunstall, Lewis and Von Werra, Leandro and Wolf, Thomas},
  year={2022},
  publisher={" O'Reilly Media, Inc."}
}

@article{llama3modelcard,
title={Llama 3 Model Card},
author={AI@Meta},
year={2024},
url = {https://github.com/meta-llama/llama3/blob/main/MODEL_CARD.md}
}

@misc{he2021debertav3,
      title={DeBERTaV3: Improving DeBERTa using ELECTRA-Style Pre-Training with Gradient-Disentangled Embedding Sharing}, 
      author={Pengcheng He and Jianfeng Gao and Weizhu Chen},
      year={2021},
      eprint={2111.09543},
      archivePrefix={arXiv},
      primaryClass={cs.CL}
}

@article{liu2019roberta,
  title={Roberta: A robustly optimized bert pretraining approach},
  author={Liu, Yinhan and Ott, Myle and Goyal, Naman and Du, Jingfei and Joshi, Mandar and Chen, Danqi and Levy, Omer and Lewis, Mike and Zettlemoyer, Luke and Stoyanov, Veselin},
  journal={arXiv preprint arXiv:1907.11692},
  year={2019}
}

@article{yang2019xlnet,
  title={Xlnet: Generalized autoregressive pretraining for language understanding},
  author={Yang, Zhilin and Dai, Zihang and Yang, Yiming and Carbonell, Jaime and Salakhutdinov, Russ R and Le, Quoc V},
  journal={Advances in neural information processing systems},
  volume={32},
  year={2019}
}

@article{clark2020electra,
  title={Electra: Pre-training text encoders as discriminators rather than generators},
  author={Clark, Kevin and Luong, Minh-Thang and Le, Quoc V and Manning, Christopher D},
  journal={arXiv preprint arXiv:2003.10555},
  year={2020}
}

@article{brown2020language,
  title={Language models are few-shot learners},
  author={Brown, Tom and Mann, Benjamin and Ryder, Nick and Subbiah, Melanie and Kaplan, Jared D and Dhariwal, Prafulla and Neelakantan, Arvind and Shyam, Pranav and Sastry, Girish and Askell, Amanda and others},
  journal={Advances in neural information processing systems},
  volume={33},
  pages={1877--1901},
  year={2020}
}

@inproceedings{kwon2023efficient,
  title={Efficient Memory Management for Large Language Model Serving with PagedAttention},
  author={Woosuk Kwon and Zhuohan Li and Siyuan Zhuang and Ying Sheng and Lianmin Zheng and Cody Hao Yu and Joseph E. Gonzalez and Hao Zhang and Ion Stoica},
  booktitle={Proceedings of the ACM SIGOPS 29th Symposium on Operating Systems Principles},
  year={2023}
}

@article{BasilottaGmezPablos2022,
  doi = {10.1186/s41239-021-00312-8},
  year = {2022},
  month = feb,
  publisher = {Springer Science and Business Media {LLC}},
  volume = {19},
  number = {1},
  author = {Ver{\'{o}}nica Basilotta-G{\'{o}}mez-Pablos and Mar{\'{\i}}a Matarranz and Luis-Alberto Casado-Aranda and Ana Otto},
  title = {Teachers' digital competencies in higher education: a systematic literature review},
  journal = {International Journal of Educational Technology in Higher Education}
}

@article{Reich2022,
	author = { Justin Reich },
	title = { Teaching Drills: Advancing Practice-Based Teacher Education through Short, Low-Stakes, High-Frequency Practice },
	journal = { Journal of Technology and Teacher Education },
	volume = { 30 },
	number = { 2 },
	year = { 2022 },
	month = { August },
	pages = { 217--228 },
	address = { Waynesville, NC USA },
	publisher = { Society for Information Technology & Teacher Education },
	issn = { 1059-7069 },
}

@inproceedings{wolf-etal-2020-transformers,
    title = "Transformers: State-of-the-Art Natural Language Processing",
    author = "Thomas Wolf and Lysandre Debut and Victor Sanh and Julien Chaumond and Clement Delangue and Anthony Moi and Pierric Cistac and Tim Rault and Rémi Louf and Morgan Funtowicz and Joe Davison and Sam Shleifer and Patrick von Platen and Clara Ma and Yacine Jernite and Julien Plu and Canwen Xu and Teven Le Scao and Sylvain Gugger and Mariama Drame and Quentin Lhoest and Alexander M. Rush",
    booktitle = "Proceedings of the 2020 Conference on Empirical Methods in Natural Language Processing: System Demonstrations",
    month = oct,
    year = "2020",
    address = "Online",
    publisher = "Association for Computational Linguistics",
    doi = "10.18653/v1/2020.emnlp-demos.6",
    pages = "38--45"
}

@misc{dettmers2023qlora,
      title={QLoRA: Efficient Finetuning of Quantized {LLMs}}, 
      author={Tim Dettmers and Artidoro Pagnoni and Ari Holtzman and Luke Zettlemoyer},
      year={2023},
      eprint={2305.14314},
      archivePrefix={arXiv},
      primaryClass={cs.LG}
}

@misc{hu2021lora,
      title={LoRA: Low-Rank Adaptation of Large Language Models}, 
      author={Edward J. Hu and Yelong Shen and Phillip Wallis and Zeyuan Allen-Zhu and Yuanzhi Li and Shean Wang and Lu Wang and Weizhu Chen},
      year={2021},
      eprint={2106.09685},
      archivePrefix={arXiv},
      primaryClass={cs.CL}
}

@inproceedings{dettmers2023case,
  title={The case for 4-bit precision: k-bit inference scaling laws},
  author={Dettmers, Tim and Zettlemoyer, Luke},
  booktitle={International Conference on Machine Learning},
  pages={7750--7774},
  year={2023},
  organization={PMLR}
}

@article{MAIER2022100080,
title = {Personalized feedback in digital learning environments: Classification framework and literature review},
journal = {Computers and Education: Artificial Intelligence},
volume = {3},
pages = {100080},
year = {2022},
issn = {2666-920X},
doi = {10.1016/j.caeai.2022.100080},
author = {Uwe Maier and Christian Klotz},
}

@article{Ledger2022,
  doi = {10.1016/j.compedu.2021.104385},
  year = {2022},
  month = mar,
  publisher = {Elsevier {BV}},
  volume = {178},
  pages = {104385},
  author = {Susan Ledger and Madeline Burgess and Natasha Rappa and Brad Power and Kok Wai Wong and Timothy Teo and Bruce Hilliard},
  title = {Simulation platforms in initial teacher education: Past practice informing future potentiality},
  journal = {Computers \& Education}
}

@article{min2023recent,
  title={Recent advances in natural language processing via large pre-trained language models: A survey},
  author={Min, Bonan and Ross, Hayley and Sulem, Elior and Veyseh, Amir Pouran Ben and Nguyen, Thien Huu and Sainz, Oscar and Agirre, Eneko and Heintz, Ilana and Roth, Dan},
  journal={ACM Computing Surveys},
  volume={56},
  number={2},
  pages={1--40},
  year={2023},
  publisher={ACM New York, NY},
  doi={10.1145/3605943}
}

@article{Chang2024,
  title = {A Survey on Evaluation of Large Language Models},
  volume = {15},
  ISSN = {2157-6912},
  DOI = {10.1145/3641289},
  number = {3},
  journal = {ACM Transactions on Intelligent Systems and Technology},
  publisher = {Association for Computing Machinery (ACM)},
  author = {Chang,  Yupeng and Wang,  Xu and Wang,  Jindong and Wu,  Yuan and Yang,  Linyi and Zhu,  Kaijie and Chen,  Hao and Yi,  Xiaoyuan and Wang,  Cunxiang and Wang,  Yidong and Ye,  Wei and Zhang,  Yue and Chang,  Yi and Yu,  Philip S. and Yang,  Qiang and Xie,  Xing},
  year = {2024},
  month = mar,
  pages = {1–45}
}

@inbook{Kurni2023,
  title = {Intelligent Tutoring Systems},
  ISBN = {9783031326530},
  DOI = {10.1007/978-3-031-32653-0_2},
  booktitle = {A Beginner’s Guide to Introduce Artificial Intelligence in Teaching and Learning},
  publisher = {Springer International Publishing},
  author = {Kurni,  Muralidhar and Mohammed,  Mujeeb Shaik and Srinivasa,  K G},
  year = {2023},
  pages = {29–44}
}

@inbook{LittenbergTobias2022,
  title = {Comparing Few-Shot Learning with GPT-3 to Traditional Machine Learning Approaches for Classifying Teacher Simulation Responses},
  ISBN = {9783031116476},
  ISSN = {1611-3349},
  DOI = {10.1007/978-3-031-11647-6_95},
  booktitle = {Lecture Notes in Computer Science},
  publisher = {Springer International Publishing},
  author = {Littenberg-Tobias,  Joshua and Marvez,  G. R. and Hillaire,  Garron and Reich,  Justin},
  year = {2022},
  pages = {471–474}
}

@article{Naser2021,
  title = {Error Metrics and Performance Fitness Indicators for Artificial Intelligence and Machine Learning in Engineering and Sciences},
  volume = {3},
  ISSN = {2730-9894},
  DOI = {10.1007/s44150-021-00015-8},
  number = {4},
  journal = {Architecture,  Structures and Construction},
  publisher = {Springer Science and Business Media LLC},
  author = {Naser,  M. Z. and Alavi,  Amir H.},
  year = {2021},
  month = nov,
  pages = {499–517}
}

@article{Yan2023,
  title = {Practical and ethical challenges of large language models in education: A systematic scoping review},
  volume = {55},
  ISSN = {1467-8535},
  DOI = {10.1111/bjet.13370},
  number = {1},
  journal = {British Journal of Educational Technology},
  publisher = {Wiley},
  author = {Yan,  Lixiang and Sha,  Lele and Zhao,  Linxuan and Li,  Yuheng and Martinez‐Maldonado,  Roberto and Chen,  Guanliang and Li,  Xinyu and Jin,  Yueqiao and Gašević,  Dragan},
  year = {2023},
  month = aug,
  pages = {90–112}
}

@inbook{Shreyashree2022,
  title = {A Literature Review on Bidirectional Encoder Representations from Transformers},
  ISBN = {9789811667237},
  ISSN = {2367-3389},
  DOI = {10.1007/978-981-16-6723-7_23},
  booktitle = {Lecture Notes in Networks and Systems},
  publisher = {Springer Nature Singapore},
  author = {Shreyashree,  S. and Sunagar,  Pramod and Rajarajeswari,  S. and Kanavalli,  Anita},
  year = {2022},
  pages = {305–320}
}

@ARTICLE{Yenduri2024,
  author={Yenduri, Gokul and Ramalingam, M. and Selvi, G. Chemmalar and Supriya, Y. and Srivastava, Gautam and Maddikunta, Praveen Kumar Reddy and Raj, G. Deepti and Jhaveri, Rutvij H. and Prabadevi, B. and Wang, Weizheng and Vasilakos, Athanasios V. and Gadekallu, Thippa Reddy},
  journal={IEEE Access}, 
  title={{GPT (Generative Pre-Trained Transformer)}— {A} Comprehensive Review on Enabling Technologies, Potential Applications, Emerging Challenges, and Future Directions}, 
  year={2024},
  volume={12},
  number={},
  pages={54608-54649},
  doi={10.1109/ACCESS.2024.3389497}}

@article{Patwardhan2023,
  title = {Transformers in the Real World: A Survey on {NLP} Applications},
  volume = {14},
  ISSN = {2078-2489},
  DOI = {10.3390/info14040242},
  number = {4},
  journal = {Information},
  publisher = {MDPI AG},
  author = {Patwardhan,  Narendra and Marrone,  Stefano and Sansone,  Carlo},
  year = {2023},
  month = apr,
  pages = {242}
}

@article{Abdalrazaq2023,
  title = {Large Language Models in Medical Education: Opportunities,  Challenges,  and Future Directions},
  volume = {9},
  ISSN = {2369-3762},
  DOI = {10.2196/48291},
  journal = {JMIR Medical Education},
  publisher = {JMIR Publications Inc.},
  author = {Abd-alrazaq,  Alaa and AlSaad,  Rawan and Alhuwail,  Dari and Ahmed,  Arfan and Healy,  Padraig Mark and Latifi,  Syed and Aziz,  Sarah and Damseh,  Rafat and Alabed Alrazak,  Sadam and Sheikh,  Javaid},
  year = {2023},
  month = jun,
  pages = {e48291}
}

@article{mizumoto2023exploring,
  title={Exploring the potential of using an {AI} language model for automated essay scoring},
  author={Mizumoto, Atsushi and Eguchi, Masaki},
  journal={Research Methods in Applied Linguistics},
  volume={2},
  number={2},
  pages={100050},
  year={2023},
  publisher={Elsevier}
}

@inproceedings{Dai2023,
  title = {Can Large Language Models Provide Feedback to Students? {A} Case Study on {ChatGPT}},
  DOI = {10.1109/icalt58122.2023.00100},
  booktitle = {2023 IEEE International Conference on Advanced Learning Technologies (ICALT)},
  publisher = {IEEE},
  author = {Dai,  Wei and Lin,  Jionghao and Jin,  Hua and Li,  Tongguang and Tsai,  Yi-Shan and Gašević,  Dragan and Chen,  Guanliang},
  year = {2023},
  month = JUL,
  pages = {323-325}
}

@article{DEEVA2021104094,
title = {A review of automated feedback systems for learners: Classification framework, challenges and opportunities},
journal = {Computers \& Education},
volume = {162},
pages = {104094},
year = {2021},
issn = {0360-1315},
doi = {10.1016/j.compedu.2020.104094},
author = {Galina Deeva and Daria Bogdanova and Estefanía Serral and Monique Snoeck and Jochen {De Weerdt}},
}

@article{gombert2023coding,
  title={Coding energy knowledge in constructed responses with explainable {NLP} models},
  author={Gombert, Sebastian and Di Mitri, Daniele and Karademir, Onur and Kubsch, Marcus and Kolbe, Hannah and Tautz, Simon and Grimm, Adrian and Bohm, Isabell and Neumann, Knut and Drachsler, Hendrik},
  journal={Journal of Computer Assisted Learning},
  volume={39},
  number={3},
  pages={767--786},
  year={2023},
  publisher={Wiley Online Library},
  doi={10.1111/jcal.12767}
}

@Article{app11010095,
AUTHOR = {García-Molina, Sergio and Alario-Hoyos, Carlos and Moreno-Marcos, Pedro Manuel and Muñoz-Merino, Pedro J. and Estévez-Ayres, Iria and Delgado Kloos, Carlos},
TITLE = {An Algorithm and a Tool for the Automatic Grading of {MOOC} Learners from Their Contributions in the Discussion Forum},
JOURNAL = {Applied Sciences},
VOLUME = {11},
YEAR = {2021},
NUMBER = {1},
ARTICLE-NUMBER = {95},
ISSN = {2076-3417},
DOI = {10.3390/app11010095}
}

@article{Ochoa2020,
   author = {Xavier Ochoa and Federico Dominguez},
   doi = {10.1111/bjet.12987},
   issn = {0007-1013},
   issue = {5},
   journal = {British Journal of Educational Technology},
   month = {9},
   pages = {1615-1630},
   title = {Controlled evaluation of a multimodal system to improve oral presentation skills in a real learning setting},
   volume = {51},
   year = {2020},
}

@article{Sallam2023,
  title = {The Utility of ChatGPT as an Example of Large Language Models in Healthcare Education,  Research and Practice: Systematic Review on the Future Perspectives and Potential Limitations},
  DOI = {10.1101/2023.02.19.23286155},
  publisher = {Cold Spring Harbor Laboratory},
  author = {Sallam,  Malik},
  year = {2023},
  month = feb,
  journal = {medRxiv}
}

@article{cavalvanti2021,
title = {Automatic feedback in online learning environments: A systematic literature review},
journal = {Computers and Education: Artificial Intelligence},
volume = {2},
pages = {100027},
year = {2021},
issn = {2666-920X},
doi = {10.1016/j.caeai.2021.100027},
author = {Anderson Pinheiro Cavalcanti and Arthur Barbosa and Ruan Carvalho and Fred Freitas and Yi-Shan Tsai and Dragan Gašević and Rafael Ferreira Mello},
}

@article{Sharma2025,
  title = {The role of large language models in personalized learning: a systematic review of educational impact},
  volume = {6},
  ISSN = {2662-9984},
  DOI = {10.1007/s43621-025-01094-z},
  number = {1},
  journal = {Discover Sustainability},
  publisher = {Springer Science and Business Media LLC},
  author = {Sharma,  Sahil and Mittal,  Puneet and Kumar,  Mukesh and Bhardwaj,  Vivek},
  year = {2025},
  month = apr 
}

@inproceedings{Wang2025,
  series = {CHI EA ’25},
  title = {LearnMate: Enhancing Online Education with {LLM}-Powered Personalized Learning Plans and Support},
  DOI = {10.1145/3706599.3719857},
  booktitle = {Proceedings of the Extended Abstracts of the CHI Conference on Human Factors in Computing Systems},
  publisher = {ACM},
  author = {Wang,  Xinyu Jessica and Lee,  Christine P. and Mutlu,  Bilge},
  year = {2025},
  month = apr,
  pages = {1–10},
  collection = {CHI EA ’25}
}

@inproceedings{Chen2024,
  series = {L@S ’24},
  title = {{GPTutor}: Great Personalized Tutor with Large Language Models for Personalized Learning Content Generation},
  DOI = {10.1145/3657604.3664718},
  booktitle = {Proceedings of the Eleventh ACM Conference on Learning @ Scale},
  publisher = {ACM},
  author = {Chen,  Eason and Lee,  Jia-En and Lin,  Jionghao and Koedinger,  Kenneth},
  year = {2024},
  month = jul,
  pages = {539–541},
  collection = {L@S ’24}
}

@inbook{Hamdi2025,
  title = {{LLM-SEM}: A Sentiment-Based Student Engagement Metric Using {LLMS} for E-Learning Platforms},
  ISBN = {9783031913549},
  ISSN = {2367-4520},
  DOI = {10.1007/978-3-031-91354-9_12},
  booktitle = {Advances on Intelligent Computing and Data Science II},
  publisher = {Springer Nature Switzerland},
  author = {Hamdi,  Ali and Mazrou,  Ahmed Abdelmoneim and Shaltout,  Mohamed},
  year = {2025},
  pages = {145–154}
}

@article{Mendona2025,
  title = {Evaluating {LLMs} for Automated Scoring in Formative Assessments},
  volume = {15},
  ISSN = {2076-3417},
  DOI = {10.3390/app15052787},
  number = {5},
  journal = {Applied Sciences},
  publisher = {MDPI AG},
  author = {Mendon\c{c}a,  Pedro C. and Quintal,  Filipe and Mendon\c{c}a,  Fábio},
  year = {2025},
  month = mar,
  pages = {2787}
}

@INPROCEEDINGS{10673924,
  author={Lee, Jung X. and Song, Yeong-Tae},
  booktitle={2024 IEEE/ACIS 27th International Conference on Software Engineering, Artificial Intelligence, Networking and Parallel/Distributed Computing (SNPD)}, 
  title={College Exam Grader using {LLM AI} models}, 
  year={2024},
  volume={},
  number={},
  pages={282-289},
  doi={10.1109/SNPD61259.2024.10673924}}

@article{Morris2024,
  title = {Automated Scoring of Constructed Response Items in Math Assessment Using Large Language Models},
  volume = {35},
  ISSN = {1560-4306},
  DOI = {10.1007/s40593-024-00418-w},
  number = {2},
  journal = {International Journal of Artificial Intelligence in Education},
  publisher = {Springer Science and Business Media LLC},
  author = {Morris,  Wesley and Holmes,  Langdon and Choi,  Joon Suh and Crossley,  Scott},
  year = {2024},
  month = jul,
  pages = {559–586}
}

@inproceedings{Shaikh2023,
  title = {Exploring the potential of {Large-Language Models (LLMs)} for student feedback sentiment analysis},
  DOI = {10.1109/fit60620.2023.00047},
  booktitle = {2023 International Conference on Frontiers of Information Technology (FIT)},
  publisher = {IEEE},
  author = {Shaikh,  Sarang and Daudpota,  Sher Muhammad and Yayilgan,  Sule Yildirim and Sindhu,  Sindhu},
  year = {2023},
  month = dec,
  pages = {214–219}
}

@inbook{Bhowmik2024,
  title = {Evaluation of an {LLM}-Powered Student Agent for Teacher Training},
  ISBN = {9783031723124},
  ISSN = {1611-3349},
  DOI = {10.1007/978-3-031-72312-4_7},
  booktitle = {Technology Enhanced Learning for Inclusive and Equitable Quality Education},
  publisher = {Springer Nature Switzerland},
  author = {Bhowmik,  Saptarshi and West,  Luke and Barrett,  Alex and Zhang,  Nuodi and Dai,  Chih-Pu and Sokolikj,  Zlatko and Southerland,  Sherry and Yuan,  Xin and Ke,  Fengfeng},
  year = {2024},
  pages = {68–74}
}

@article{Hu2025,
  title = {Exploring the potential of {LLM} to enhance teaching plans through teaching simulation},
  volume = {10},
  ISSN = {2056-7936},
  DOI = {10.1038/s41539-025-00300-x},
  number = {1},
  journal = {{NPJ} Science of Learning},
  publisher = {Springer Science and Business Media LLC},
  author = {Hu,  Bihao and Zhu,  Jiayi and Pei,  Yiying and Gu,  Xiaoqing},
  year = {2025},
  month = feb 
}

@article{Zheng2025,
  title = {Teaching via LLM-enhanced simulations: Authenticity and barriers to suspension of disbelief},
  volume = {65},
  ISSN = {1096-7516},
  DOI = {10.1016/j.iheduc.2024.100990},
  journal = {The Internet and Higher Education},
  publisher = {Elsevier BV},
  author = {Zheng,  Longwei and Jiang,  Fei and Gu,  Xiaoqing and Li,  Yuanyuan and Wang,  Gong and Zhang,  Haomin},
  year = {2025},
  month = apr,
  pages = {100990}
}

\section*{Acknowledgment}
\addtxt{The first author acknowledges the support provided by the FPU program of the University of Alcalá and the Spanish Ministry of Science, Innovation and Universities (FPU23/01120). The second author thanks grants PID2021-122466OB-I00 and PRE2022-102391 (MCIN/AEI/10.13039/501100011033/FEDER, Spain), which supported his PhD studies and a research stay at the Massachusetts Institute of Technology. The third and fourth authors acknowledge the Mobility Program of the University of Alcalá. All authors from the University of Alcalá also acknowledge the project ``Using Natural Language Processing to Create Automatic Personalized Feedback in Teacher Simulations,'' funded by the MISTI Global Seed Funds program, which supported the collaboration between the Massachusetts Institute of Technology and the University of Alcalá.}

\section*{Data availability statement}
The data and source code will be made available on request. For requesting data, please write to the corresponding author.

\section*{Ethics statement}
This study was reviewed and deemed exempt from ethics approval by the Massachusetts Institute of Technology Committee on the Use of Humans as Experimental Subjects with the reference number: E-1872 (Designing for Transformative Equitable Teaching Practice Spaces and Mindsets), dated January 10, 2020. The first vignette of TM simulations provides a consent agreement to either allow or decline the use of the participant's data for research purposes. It also indicates that all data are protected under the Massachusetts Institute of Technology's IRB review procedures.

\section*{Declaration of competing interest}
The authors declare no competing interests.

\section*{\addtxt{CRediT  authorship contribution statement}}
\addtxt{
\textbf{David de-Fitero-Dominguez:} Conceptualization, Investigation, Software, Visualization, Validation, Writing - Original Draft, Writing - Review \& Editing. 
\textbf{Mariano Albaladejo-González:} Conceptualization, Investigation, Resources, Data Curation, Validation, Writing - Original Draft, Writing - Review \& Editing. 
\textbf{Antonio Garcia-Cabot:} Conceptualization, Investigation, Methodology, Resources, Visualization, Validation, Supervision, Writing - Original Draft, Writing - Review \& Editing. 
\textbf{Eva Garcia-Lopez:} Conceptualization, Investigation, Methodology, Resources, Validation, Visualization, Supervision, Writing - Original Draft, Writing - Review \& Editing. 
\textbf{Antonio Moreno-Cediel:} Conceptualization, Investigation, Software, Validation, Writing - Review \& Editing. 
\textbf{Erin Barno:} Conceptualization, Investigation, Resources, Data Curation, Validation, Writing - Review \& Editing. 
\textbf{Justin Reich:} Conceptualization, Investigation, Resources, Validation, Supervision, Project Administration, Writing - Review \& Editing.
}

\newpage

\appendix

\section{Structured Reasoning Examples}
\label{appendix:cot_examples}

This section provides complete reasoning traces from Qwen 3's structured reasoning experiments, demonstrating how the model approaches educational characteristic identification tasks.

\subsection{Example 1}
\label{appendix:cot_example1}

\textbf{Teacher Response:} ``Probably not the whole quiz. He did understand hyperbole so I might have him take parts of the quiz and then exempt the rest of it."

\textbf{Characteristic:} modification (Label: 1, Prediction: 1)

\begin{lstlisting}[caption=Structured reasoning for modification characteristic, label=listing:cot_example1]
Okay, let's tackle this evaluation. The user wants to know if the novice
teacher's response suggests a modification to a scheduled quiz. The response
given is: "Probably not the whole quiz. He did understand hyperbole so I might
have him take parts of the quiz and then exempt the rest of it."

First, I need to parse the key elements here. The teacher is talking about a
quiz. The student didn't do well, but the teacher thinks the student understood
hyperbole. So, instead of giving the whole quiz, the teacher is considering
having the student take parts of it. The phrase "exempt the rest of it" is
important. Exempting parts of the quiz would mean those parts aren't counted
towards the grade.

Now, the characteristic we're looking for is any modification to the scheduled
quiz, like moving the date, allowing a retake, or not counting the grade. The
teacher's response doesn't mention changing the date or allowing a retake.
However, they are suggesting not to count the entire quiz. By exempting the
rest, they're modifying the quiz's impact on the student's grade.

Wait, does "exempt the rest of it" mean the student doesn't have to do those
parts, or that those parts aren't graded? If the student takes parts of the
quiz and the rest is exempt, it could mean those parts aren't counted. That
would be a modification of the quiz's grading. So, even though the teacher
isn't changing the date or offering a retake, they are altering the quiz's
structure or grading.

But the characteristic specifically lists "not counting the grade" as a
possible modification. If the teacher is exempting parts, that's not counting
those parts towards the grade. So that would fit under the characteristic.
Therefore, the response does suggest a modification to the scheduled quiz by
altering which parts are graded.

I need to make sure there's no ambiguity. The teacher is modifying the quiz by
exempting parts, which is a change to the original plan. So the characteristic
is present.
\end{lstlisting}

\subsection{Example 2}
\label{appendix:cot_example2}

\textbf{Teacher Response:} ``Hey, Jeremy. We missed you yesterday. Are you feeling better today? (Let him answer.) (Assuming it's yes since he's back today, I'd continue): Yesterday, we made some changes to the mentor sentence. I'd like you to join the small group right now. We're going to review yesterday's lesson again before starting the new one to make sure everyone understands. You'll be able to see what you missed, be part of the discussion, and ask any questions you might have. Does that sound okay?"

\textbf{Characteristic:} teacher\_catch\_up (Label: 0, Prediction: 1)

\begin{lstlisting}[caption=Structured reasoning for teacher catch-up characteristic, label=listing:cot_example2]
Okay, let me try to work through this. The task is to determine if the novice
teacher's response includes the characteristic of "offers to meet with Jeremy
to help him catch up on missed classwork."

First, I need to understand the scenario. The novice teacher is in a simulation
where they're pretending to be a teacher. The specific situation is that Jeremy
was absent the previous day, and the teacher is now addressing that. The
response starts with the teacher greeting Jeremy and asking if he's feeling
better, which is a good sign of checking in. Then, the teacher assumes Jeremy
is okay because he's back today and proceeds to explain that they made changes
to the mentor sentence. The teacher invites Jeremy to join the small group to
review the previous lesson before starting the new one, allowing Jeremy to see
what he missed, participate in the discussion, and ask questions.

Now, the key characteristic to check is whether the teacher offers to meet with
Jeremy to help him catch up. The teacher's plan is to have Jeremy join the
small group for a review session. This seems like an opportunity for Jeremy to
catch up on the missed work. The teacher is providing a structured way for
Jeremy to re-engage with the material, which implies a meeting or session to
address the missed classwork.

But does "joining the small group to review" count as offering to meet? The
teacher is not explicitly saying they'll meet one-on-one, but rather
integrating Jeremy into a group activity that's meant to help him catch up. The
teacher mentions that Jeremy will be able to see what he missed, participate,
and ask questions. This suggests that the teacher is facilitating a catch-up
session as part of the group work.

However, the characteristic specifically mentions "offers to meet," which might
imply a more direct or individualized meeting. The teacher's approach here is
more about including Jeremy in a group review rather than a dedicated meeting.
But in a classroom setting, participating in a small group to review the lesson
could be considered a form of meeting or catching up. The teacher is making an
effort to ensure Jeremy understands the material he missed, which aligns with
the intent of the characteristic.

Another angle: the teacher is not explicitly offering a separate meeting but is
instead incorporating Jeremy into the lesson review. The question is whether
this counts as "offering to meet." If the characteristic is about providing an
opportunity to catch up, then yes. But if it's about a specific meeting
arrangement, then maybe not. However, the teacher's response does include the
plan to review the lesson, which is a way to help Jeremy catch up.

I think the key here is that the teacher is taking proactive steps to help
Jeremy catch up by including him in the review. So even though it's a group
activity, the intent is to help him catch up. Therefore, the characteristic is
present.
\end{lstlisting}

\subsection{Example 3}
\label{appendix:cot_example3}

\textbf{Teacher Response:} ``I want to check in with Jeremy privately to see what he does not understand and try to help him out. Also I would take some of that time to make sure that he is okay and how are things at home."

\textbf{Characteristic:} change\_for (Label: 1, Prediction: 0)

\begin{lstlisting}[caption=Structured reasoning for instructional change characteristic, label=listing:cot_example3]
Okay, let's break this down. The task is to determine if the novice teacher's
response shows the characteristic of "Proposes changes to the planned
instruction."

First, I need to understand what that characteristic means. Proposing changes
to planned instruction would involve adjusting the original lesson plan or
teaching strategy based on student needs or observations. The teacher might
modify activities, pacing, materials, or assessment methods.

Now, looking at the response: "I want to check in with Jeremy privately to see
what he does not understand and try to help him out. Also I would take some of
that time to make sure that he is okay and how are things at home."

The teacher is planning to check in with Jeremy individually. This seems like a
form of differentiated instruction or individualized support. However, the key
here is whether this is a change to the planned instruction. The original plan
might have been a standard lesson, and the teacher is now deviating to provide
extra help.

But wait, the response doesn't mention changing the lesson plan itself.
Instead, it's about addressing Jeremy's needs through private check-ins. That
could be part of formative assessment or differentiated instruction, which are
part of adjusting instruction. However, the term "proposes changes to the
planned instruction" might specifically refer to altering the lesson's
structure, content, or activities.

The teacher is using time outside the original plan (checking in privately) to
support Jeremy. This could be considered an adjustment to the instructional
approach. However, the response doesn't explicitly mention changing the lesson
plan's content or structure. It's more about addressing individual needs
through additional support rather than modifying the planned instruction.

Another angle: if the teacher was supposed to follow a specific lesson plan,
and now they're diverting time to check in, that might be a change. But the
original instruction might have included time for individual check-ins. The
response doesn't clarify if this is a deviation from the planned schedule.

So, the answer hinges on whether the check-in is a change to the planned
instruction. If the original plan didn't include this, then yes, it's a change.
If it was part of the plan, then maybe not. Since the response doesn't specify
the original plan, we have to assume that checking in privately is an
adjustment. However, the characteristic is about proposing changes, not just
adjusting.

The teacher is suggesting a specific action (checking in) which could be part
of modifying instruction. However, the phrasing "proposes changes" might
require more explicit mention of altering the lesson plan. The response is more
about addressing student needs rather than changing the instructional plan
itself.

Therefore, the characteristic might not be present because the focus is on
individual support rather than changing the planned instruction. The answer
would be 0.
\end{lstlisting}

\section{Prompt design}
\label{appendix:prompt_design}
\subsection{Zero-shot Prompt}
\label{appendix:zero_shot_prompt}

For the zero-shot configuration, the following prompt was used.

\begin{lstlisting}[caption=Zero-shot configuration prompt, label=listing:zero_shot_prompt]
<|begin_of_text|><|start_header_id|>system<|end_header_id|>

You are a teacher educator who is having their novice teachers go through a practice experience pretending to be teachers in a specific scenario that is in a simulation. Output a '0' if the characteristic is not present in the teacher's response and a '1' if it is.<|eot_id|><|start_header_id|>user<|end_header_id|>

You, the teacher educator, want to evaluate whether the novice teacher's response exhibits a specific characteristic. Here is the novice teacher's response:

Thanks for this Jeremy! I'm sorry to hear you weren't feeling well. Could you stay after class for a quick 5 minutes so that I can run-through the work with you?

This is the characteristic you are looking for in the response:

Suggests that Jeremy needs to catch up on the academic work.

You have to evaluate if the characteristic is present in the teacher's response.<|eot_id|><|start_header_id|>assistant<|end_header_id|> 
\end{lstlisting}

\subsection{Few-shot Prompt}
\label{appendix:few_shot_prompt}
In the few-shot configuration, the following prompt was used to guide the Llama 3 model in classifying whether a specific characteristic was present in the teacher candidate's response. The prompt includes five examples sampled from the training subset to help the model understand the task. Note that the same examples are used for all three experiments, however, the characteristics in these examples are not included in the test sets of any of the experiments.

\begin{lstlisting}[caption=Few-shot configuration prompt, label=listing:few_shot_prompt]
<|begin_of_text|><|start_header_id|>system<|end_header_id|>

You are a teacher educator who is having their novice teachers go through a practice experience pretending to be teachers in a specific scenario that is in a simulation. Output a '0' if the characteristic is not present in the teacher's response and a '1' if it is.<|eot_id|><|start_header_id|>user<|end_header_id|>

You, the teacher educator, want to evaluate whether the novice teacher's response exhibits a specific characteristic. Here is the novice teacher's response:

Thanks for this Jeremy! I'm sorry to hear you weren't feeling well. Could you stay after class for a quick 5 minutes so that I can run-through the work with you?

This is the characteristic you are looking for in the response:

Suggests that Jeremy needs to catch up on the academic work.

You have to evaluate if the characteristic is present in the teacher's response.<|eot_id|><|start_header_id|>assistant<|end_header_id|>

0<|eot_id|><|start_header_id|>user<|end_header_id|>

You, the teacher educator, want to evaluate whether the novice teacher's response exhibits a specific characteristic. Here is the novice teacher's response:

He is not prepared to face the quiz as he has not completed the formatives.

This is the characteristic you are looking for in the response:

Observes or states that Jeremy isn't feeling well.

You have to evaluate if the characteristic is present in the teacher's response.<|eot_id|><|start_header_id|>assistant<|end_header_id|>

0<|eot_id|><|start_header_id|>user<|end_header_id|>

You, the teacher educator, want to evaluate whether the novice teacher's response exhibits a specific characteristic. Here is the novice teacher's response:

To be honest, I want to both review the lesson in a small group AND check in privately with Jeremy (but I can't select both). I would review the lesson in a small group with the 7 who had an error. For Jeremy, I can completely understand why he wrote what he did in his journal - he wasn't there for direct instruction the day before so he's probably feeling a bit lost. I would either make time in class for me to pull Jeremy aside to catch him up, but ideally, I would like to meet with him during lunch or before/after school to help him.

This is the characteristic you are looking for in the response:

Identifies an academic challenge Jeremy faces.

You have to evaluate if the characteristic is present in the teacher's response.<|eot_id|><|start_header_id|>assistant<|end_header_id|>

1<|eot_id|><|start_header_id|>user<|end_header_id|>

You, the teacher educator, want to evaluate whether the novice teacher's response exhibits a specific characteristic. Here is the novice teacher's response:

Yes. He needs to understand what the expectations are for him and the class. If he does well, great. If he does not do well, he can redo some work so he understands the content better.

This is the characteristic you are looking for in the response:

States that Jeremy should take the quiz

You have to evaluate if the characteristic is present in the teacher's response.<|eot_id|><|start_header_id|>assistant<|end_header_id|>

0<|eot_id|><|start_header_id|>user<|end_header_id|>

You, the teacher educator, want to evaluate whether the novice teacher's response exhibits a specific characteristic. Here is the novice teacher's response:

I would be more insistent this time to know how is Jeremy feeling? I would ask him if he understood the topics that we've been studying. I would be asking him how i could be of further assistance to him should he not understand the topic. I'll encourage him to at least use a sentence to the topics we've gone through to know had he learned from it.

This is the characteristic you are looking for in the response:

Argues for giving more academic support to Jeremy or other students who need it.

You have to evaluate if the characteristic is present in the teacher's response.<|eot_id|><|start_header_id|>assistant<|end_header_id|>

1<|eot_id|><|start_header_id|>user<|end_header_id|>

You, the teacher educator, want to evaluate whether the novice teacher's response exhibits a specific characteristic. Here is the novice teacher's response:

{Teacher's response}

This is the characteristic you are looking for in the response:

{Characteristic description}.

You have to evaluate if the characteristic is present in the teacher's response.<|eot_id|><|start_header_id|>assistant<|end_header_id|>
\end{lstlisting}

\section{Detailed performance metrics for additional LLMs}
\label{appendix:detailed_results}

This section provides the detailed performance metrics for the additional Large Language Models (LLMs), Phi-4-mini and Qwen 3 (8B), evaluated in zero-shot and few-shot configurations without fine-tuning. The results are presented for each of the three test splits (Low-sample, Random, Performance-matched) as well as the corresponding validation splits, complementing the summary results discussed in Section \ref{sec:discussion}. Metrics include Balanced Accuracy, F1 Score, Precision, and Recall, rounded to three decimal places. The results for the non-fine-tuned Llama 3 model from the main paper experiments (`Llama 3') are also included.

\begin{table}
\begin{adjustbox}{width=1\textwidth}
\begin{tabular}{lllrrrr}
\toprule
Model & Split & Configuration & BAC & F1 Score & Precision & Recall \\
\midrule
Llama 3 & Test Low-sample & Few-shot & 0.913 & 0.884 & 0.918 & 0.854 \\
Llama 3 & Test Low-sample & Zero-shot & 0.796 & 0.740 & 0.969 & 0.599 \\
Llama 3 & Test Random & Few-shot & 0.731 & 0.671 & 0.606 & 0.752 \\
Llama 3 & Test Random & Zero-shot & 0.691 & 0.603 & 0.647 & 0.564 \\
Llama 3 & Test Performance-matched & Few-shot & 0.882 & 0.746 & 0.653 & 0.869 \\
Llama 3 & Test Performance-matched & Zero-shot & 0.742 & 0.537 & 0.445 & 0.676 \\
Llama 3 & Evaluation Low-sample & Few-shot & 0.742 & 0.530 & 0.426 & 0.702 \\
Llama 3 & Evaluation Low-sample & Zero-shot & 0.732 & 0.526 & 0.437 & 0.660 \\
Llama 3 & Evaluation Random & Few-shot & 0.731 & 0.480 & 0.372 & 0.677 \\
Llama 3 & Evaluation Random & Zero-shot & 0.752 & 0.519 & 0.419 & 0.683 \\
Llama 3 & Evaluation Performance-matched & Few-shot & 0.729 & 0.517 & 0.401 & 0.726 \\
Llama 3 & Evaluation Performance-matched & Zero-shot & 0.719 & 0.526 & 0.456 & 0.622 \\
Phi-4-mini & Test Low-sample & Few-shot & 0.874 & 0.811 & 0.801 & 0.822 \\
Phi-4-mini & Test Low-sample & Zero-shot & 0.887 & 0.854 & 0.913 & 0.803 \\
Phi-4-mini & Test Random & Few-shot & 0.830 & 0.781 & 0.692 & 0.897 \\
Phi-4-mini & Test Random & Zero-shot & 0.772 & 0.715 & 0.695 & 0.737 \\
Phi-4-mini & Test Performance-matched & Few-shot & 0.814 & 0.573 & 0.415 & 0.924 \\
Phi-4-mini & Test Performance-matched & Zero-shot & 0.860 & 0.693 & 0.579 & 0.862 \\
Phi-4-mini & Evaluation Low-sample & Few-shot & 0.784 & 0.568 & 0.438 & 0.807 \\
Phi-4-mini & Evaluation Low-sample & Zero-shot & 0.764 & 0.544 & 0.420 & 0.773 \\
Phi-4-mini & Evaluation Random & Few-shot & 0.769 & 0.514 & 0.387 & 0.767 \\
Phi-4-mini & Evaluation Random & Zero-shot & 0.791 & 0.539 & 0.405 & 0.804 \\
Phi-4-mini & Evaluation Performance-matched & Few-shot & 0.785 & 0.588 & 0.466 & 0.797 \\
Phi-4-mini & Evaluation Performance-matched & Zero-shot & 0.737 & 0.533 & 0.426 & 0.714 \\
Qwen 3 8B & Test Low-sample & Few-shot & 0.913 & 0.884 & 0.918 & 0.854 \\
Qwen 3 8B & Test Low-sample & Zero-shot & 0.876 & 0.846 & 0.938 & 0.771 \\
Qwen 3 8B & Test Low-sample & Structured reasoning few-shot & 0.879 & 0.850 & 0.938 & 0.777 \\
Qwen 3 8B & Test Low-sample & Structured reasoning zero-shot & 0.827 & 0.784 & 0.669 & 0.669 \\
Qwen 3 8B & Test Random & Few-shot & 0.776 & 0.716 & 0.781 & 0.661 \\
Qwen 3 8B & Test Random & Zero-shot & 0.750 & 0.684 & 0.712 & 0.658 \\
Qwen 3 8B & Test Random & Structured reasoning few-shot & 0.818 & 0.773 & 0.795 & 0.752 \\
Qwen 3 8B & Test Random & Structured reasoning zero-shot & 0.766 & 0.702 & 0.791 & 0.630 \\
Qwen 3 8B & Test Performance-matched & Few-shot & 0.869 & 0.765 & 0.727 & 0.807 \\
Qwen 3 8B & Test Performance-matched & Zero-shot & 0.842 & 0.688 & 0.600 & 0.807 \\
Qwen 3 8B & Test Performance-matched & Structured reasoning few-shot & 0.804 & 0.709 & 0.783 & 0.648 \\
Qwen 3 8B & Test Performance-matched & Structured reasoning zero-shot & 0.772 & 0.656 & 0.735 & 0.593 \\
Qwen 3 8B & Evaluation Low-sample & Few-shot & 0.795 & 0.629 & 0.557 & 0.723 \\
Qwen 3 8B & Evaluation Low-sample & Zero-shot & 0.755 & 0.580 & 0.528 & 0.643 \\
Qwen 3 8B & Evaluation Low-sample & Structured reasoning few-shot & 0.805 & 0.653 & 0.595 & 0.723 \\
Qwen 3 8B & Evaluation Low-sample & Structured reasoning zero-shot & 0.753 & 0.596 & 0.591 & 0.601 \\
Qwen 3 8B & Evaluation Random & Few-shot & 0.811 & 0.612 & 0.514 & 0.757 \\
Qwen 3 8B & Evaluation Random & Zero-shot & 0.755 & 0.556 & 0.498 & 0.630 \\
Qwen 3 8B & Evaluation Random & Structured reasoning few-shot & 0.801 & 0.623 & 0.556 & 0.709 \\
Qwen 3 8B & Evaluation Random & Structured reasoning zero-shot & 0.739 & 0.552 & 0.535 & 0.571 \\
Qwen 3 8B & Evaluation Performance-matched & Few-shot & 0.772 & 0.605 & 0.537 & 0.693 \\
Qwen 3 8B & Evaluation Performance-matched & Zero-shot & 0.740 & 0.577 & 0.558 & 0.598 \\
Qwen 3 8B & Evaluation Performance-matched & Structured reasoning few-shot & 0.801 & 0.654 & 0.598 & 0.722 \\
Qwen 3 8B & Evaluation Performance-matched & Structured reasoning zero-shot & 0.721 & 0.558 & 0.578 & 0.539 \\
\bottomrule
\end{tabular}
\end{adjustbox}
\caption{Detailed performance metrics including evaluation sets for Llama 3, Phi-4-mini, and Qwen 3 (8B).} 
\label{tab:appendix_detailed_results}
\end{table}




\end{document}